\documentclass[journal]{IEEEtran}
\usepackage{graphicx}
\usepackage{bm}
\usepackage{hyperref}
\usepackage{subfig}
\usepackage{multirow}
\usepackage{array}
\usepackage{epstopdf,epsfig}
\usepackage{color}
\usepackage{boxedminipage}
\usepackage[table]{xcolor}%
\usepackage{float}
\usepackage{fixltx2e} 
\usepackage{colortbl}
\usepackage{anyfontsize}
\usepackage{amsmath,amssymb} 
\usepackage[ruled]{algorithm2e}
\usepackage{changebar}
\usepackage{ulem}
\usepackage{mathrsfs}
\usepackage{multicol}
\usepackage[numbers]{natbib}
\definecolor{Gray}{gray}{0.9}
\definecolor{Wgray}{gray}{0.8}
\newcolumntype{g}{>{\columncolor{Gray}}c}
\newcolumntype{w}{>{\columncolor{Wgray}}c}

\newcommand{\Sign}[1]{\ensuremath{\operatorname*{sign}\left(#1\right)}}
\newcommand{\Tr}{\operatorname*{trace} }
\newcommand{\Diag}{\operatorname*{diag} }

\newtheorem{proposition}{Proposition}
\newenvironment{myproof}[1][Proof]{\textit{#1.} }{\ \rule{0.5em}{0.5em}}
\begin{document}

\title{
Using Curvilinear Features in Focus for Registering a Single Image to a 3D Object}

\author{Hatem A. Rashwan,
        Sylvie Chambon,
        Pierre Gurdjos, G\'{e}raldine Morin and Vincent Charvillat
\thanks{University of Toulouse, IRIT}}

\maketitle

\begin{abstract}
In the context of 2D/3D registration, this paper introduces an approach that allows to match features detected in two different modalities: photographs and 3D models, 
by using a common 2D reprensentation. 
More precisely, 2D images are matched with a set of depth images, representing the 3D model. 
After introducing the concept of curvilinear saliency, related to curvature estimation, we propose a new ridge and valley detector for depth images rendered from 3D model. 
A variant of this detector is adapted to photographs, in particular by applying it in multi-scale and by combining this feature detector with the principle of focus curves. 
Finally, a registration algorithm for determining the correct viewpoint of the 3D model and thus the pose is proposed. 
It is based on using histogram of gradients features adapted to the features manipulated in 2D and in 3D, and the introduction of repeatability scores. 
The results presented highlight the quality of the features detected, in term of repeatability, and also the interest of the approach for registration and pose estimation. 
\end{abstract}


\section{Introduction}

Many computer vision and robotic applications are used to take 2D contents as input, 
but, recently 3D contents are simultaneously available and popular. In order to benefit from both modalities, 2D/3D matching is necessary. 
For medical imaging, registration of pre-operative 3D volume data with intra-operative 2D images becomes more and more necessary to assist physicians 
in diagnosing complicated diseases easily and quickly~\cite{Markelj2012}. 
For robotic, the 2D/3D matching is very important for many tasks that need to determine the 3D pose of an object of interest: 3D navigation or object grasping~\cite{Pomerleau2015}. 
The main goal of 2D/3D registration is to find the transformation of the 3D model that defines the pose for a query 2D image.
Thus, a typical 2D/3D registration problem consists of two mutually interlocked subproblems, point correspondence and pose estimation.

To match 2D photographs directly to 3D models or points clouds, most systems rely on detecting and describing features
on both 2D/3D data and then on matching these features~\cite{Wu2008,Agarwal2009}.
Recently, some approaches are based on learning by specific supervision classifier~\cite{Krizhevsky2012,Hao2015}. 
These methods produce very interesting results, however, they require huge amount of viewpoint-annotated images to learn the classifiers.
What makes difficulty to the problem of matching 3D features of an object to 2D features of one of its photographs is that the appearance of the object dramatically depends on intrinsic characteristics of the object, like texture and color/albedo, as well as extrinsic characteristics related to the acquisition, like the camera pose and the lighting conditions.
Consequently, some approaches manually define correspondences between the query image and the 3D model, such as~\cite{Dellepiane2008}.
These manual methods can be robust but it can easily become hard to apply this manual selection to large image sets. Moreover, in this paper, we focus on automated approaches.
Note that some systems are able to generate a simultaneous acquisition of photographs and scanning of a 3D model but using this kind of systems induces limited applications.
Other methods solve the problem by distinguishing two subproblems: to choose the common representation of the data and, then, to find the correspondences. 
These methods transforms the initial 2D/3D registration problem to a 2D/2D matching problem by rendering multiple 2D images of 3D models from different viewpoints, 
such as~\cite{Campbell2001,Choy2015,Plotz2015}.

\begin{figure*}
  \centering
  \includegraphics[width=2\columnwidth]{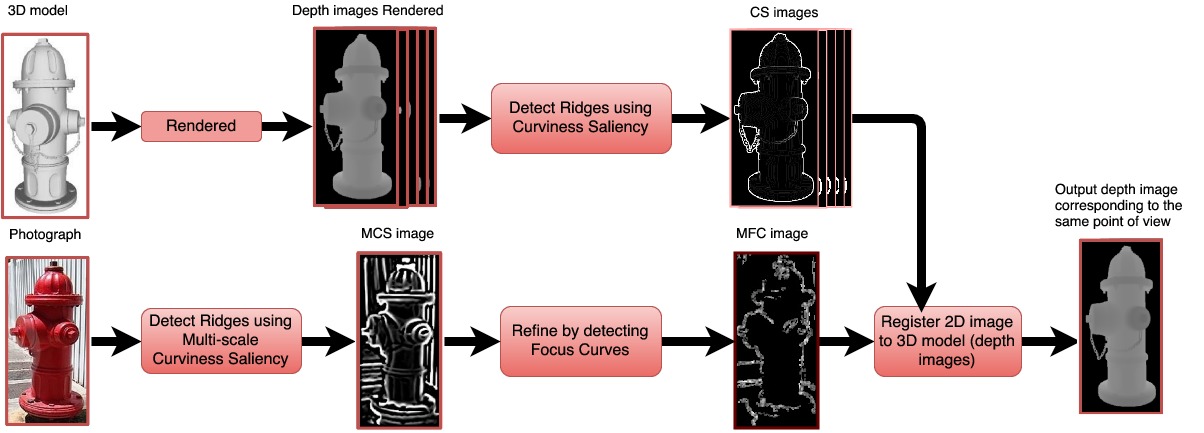}
  \caption{To compare 2D images with 3D models, we use a collection of rendered images of the 3D models from different viewpoints,
  and then we detect points of interest (curvilinear saliency) with common basis definitions between depth images and intensity images. 
  For that evaluation, each depth image is compared with the original 2D image, based on these point of interest detection, 
  and the proposed algorithm gives as output the depth image with the most similar point of view with the point of view of the 2D image. }\label{Alg}
\end{figure*}

Consequently, the first task of 2D/3D registration is to \textit{find an appropriate representation of 3D models in which reliable features can be extracted in 2D and 3D data}. 
In~\cite{Campbell2001}, synthetic images of the 3D model are rendered, while depth images are rendered in~\cite{Choy2015}. 
More recently,~\cite{Plotz2015} proposes average shading gradients. This rendering technique for a 3D model averages the gradient normals over all lighting directions 
to cope with the unknown lighting of the query image. 
The advantage of representing the 3D model by a set of depth images is that it can express the model shape independently to color and texture information.
Therefore, representing the 3D model by a set of depth images is the best option for this work, see Fig.~\ref{Alg}. 
In this case, features extracted from depth images are only related to shape information.

%

Then, the second difficulty of 2D/3D registration consists in proposing \textit{how to match entities between the two modalities in this common representation}. 
It can be partial~\cite{Irschara2009} or dense matching, based on local or global characteristics~\cite{Szeliski2002}. 
In~\cite{Campbell2001},  silhouettes extracted from synthetic images are matched to ones extracted from the color images. 
However, this method is not able to take into account most of the occluding contours that are useful for accurate pose estimation. 
In turn, in~\cite{Plotz2015}, image gradients are matched with their 3D representation. 
Since image gradients are still affected by image textures and background, this technique can fail to estimate the correct correspondences. 
A key requirement on these features, as in classic 2D matching between real images, is to be computed with a high degree of \textit{repeatability}. 
In our case, similar to the definition in~\cite{Szeliski2010}, 
the repeatability of a feature is defined as the frequency with which one detected in the depth image is found 
within $\epsilon$ pixels around the same location in the corresponding intensity image (if it is supposed that the features are not moving or are following a small deplacement).
Then, since we suppose that an individual photograph of an object of interest is acquired, in a textured environment, we will focus on comparing pre-processed features of color images
with features of a set of rendered images of 3D models, more precisely, a set of depth images, see Fig.~\ref{Alg}.

More precisely, the 3D object will be given by a set of 3D depth surfaces, which describe how the original object surface is shortened by a perspective viewing and 
the image is given by the 3D intensity surface. Since the depth and the intensity surfaces have a different order of representation, the two surfaces can not be directly matched.
Thus, bringing both rendered depth images and photographs into a common representation, such as gradient and edge representation, allows to establish a robust sparse 2D-to-3D matching~\cite{Plotz2015}.
We propose to extract gradient-based features corresponding to object's shapes in both depth and intensity images regardless of illumination and texture changes.
In other words, as 2D photographs (intensity images) are affected by background, textures and lighting changes,
we take into account these difficulties by reducing the influence of non-redundant information (i.e., color and texture) on features extracted from photographs.
It means that we extract features in depth images that highlight geometric characteristics of an object.
For photographs, we need to refine detected features by selecting salient points acquired by a camera in focus.
These points are a function of the degree of blur (blurriness) in an image.
Thus, the detected points are analyzed based on measuring the blur amount of every feature point.
Finally, what we call focus points should be able to detect the approximate shape and to discard the other components such as textures.

To summary, the contributions of this paper, as shown in Fig.~\ref{Alg}, are:
\begin{enumerate}
 \item \textbf{A ridge and valley detector, for depth images rendered from 3D model. }We name it curvilinear saliency (CS) as it is related to the curvature estimation
 (a function of the eigenvalues derived of the Hessian matrix).
 This representation directly relates to the discontinuities of the object's geometry, and, by nature, the extracted features should be robust to texture and light changes.
\item \textbf{A variant of this detector adapted to photographs. }  
This curvilinear saliency detector is applied in multi-scale by searching over all scales and all image locations
in order to identify scale-invariant interest points.
To reduce the influence of structures due to texture and background regions, we introduce the extraction of focus curvilinear saliency features.
It corresponds to ridges that are not affected by blur.
\item \textbf{A registration algorithm for determining the correct viewpoint of the 3D model and thus the pose. } 
This method is based on using histogram of gradients, HOG, features~\cite{Dalal2005}, adapted to the features manipulated in 2D and in 3D, and the introduction of 
repeatability scores.
More precisely, the HOG descriptor is computed on both depth images (i.e., curvilinear features extracted with curvilinear saliency detection) and photographs (i.e., curvilinear features in focus extracted with multi-scales curvilinear saliency detection) and it combines the curvilinear saliency value with the orientation of the curvature. The repeatability score measures the set of repeatable points detected both in a photograph and in the rendered depth images.
\end{enumerate}

After presenting the related work and reminder on differential geometry, sections~\ref{sec:relatedWork} and \ref{sec:reminder}, 
we introduce the 3D model reprensentation, section~\ref{sec:3DModelRepr}, and, then, the image representation used for 2D/3D matching, section~\ref{sec:ima:rep}. 
Then, we describe how we try to be robust to background and texture by using the same principle used in the detection of focus curves, section~\ref{sec:robustness}. 
We illustrate how this new global approach for 2D/3D matching allows to obtain more repeatable features, compared to state of the art, section~\ref{exp}. 
Finally, we explain how we obtained 2D/3D registration results, section~\ref{sec:registration}, 
and pose estimation, section~\ref{sec:pose}, by highlighting the interest of the proposed approach in these applications, before conclusion, section~\ref{sec:conclusion}. 

\section{Related work}\label{sec:relatedWork}
As mentioned earlier, a typical 2D/3D registration problem consists of two subproblems: feature correspondence and pose estimation (i.e., alignment). Thus, the related work is divided into three parts related to these subproblems: 1) detect features in 2D photography 2) detect features in a 3D model and finally 3) match 2D features to 3D features to estimate the 3D pose.

\subsection{Classic 2D feature detection}

In this section, we try to identify if, in the literature, it exists a 2D classical detector interesting in order to obtain points comparable as points detected, 
with the same principle or same tool, in 3D. 
We suppose that, for this purpose, it is necessary to detect features that are related to points of interest on the structure of the object and not on the texture or the 
light changes on the object. 
In 2D, edge detection~\cite{canny1986} based on the first-order derivative information is the initial technique. 
It can detect any kind of edges, even low contrasted edges that are not due to the structure but more due to texture. 
The second technique is to detect the interest points~\cite{harris88} that refer to point-like features in an image by analyzing the eigenvalues of the structure tensor 
(i.e., the first-order derivative) at each point -- the two eigenvalues have to be maximal to highlight a point of interest.
Again, this technique does not take into account difficulties due to textures, light changes or scale changes. 
In another way, blob detection~\cite{Smith1997,Matas2002,Lowe2004,Tuytelaars2004} provides a complementary description of image structures in terms of regions, as opposed to point-like algorithms.
These methods are based on the Hessian matrix (i.e., the second-order derivative) because of its independence to zero- and first-order changes and its good performance in computation time and accuracy.
More recently,  multi-scale approaches have been introduced, like a generalization of Harris or Laplacian detectors~\cite{Mikolajczyk2004} or
the well known approach of SIFT, Scale Invariant Features Transform~\cite{Lowe2004}.
In~\cite{Bay2008}, SURF, Speeded Up Robust Features, a detector also based on Hessian matrix analysis, is introduced to be faster than SIFT and other multi-scale techniques by using approximation of Laplacian and algorithmic tricks.
All these techniques are robust to light changes, rotations and translations. It makes the features detection invariant to view-point changes.
However, they totally rely on texture and/or intensity changes to find the features.

Curvature detection is one of the most important techniques of second-order derivative-based approaches used for extracting the structure properties. 
Recently,~\cite{Fischer2014} has proposed a detector based on curvature $\kappa$ expressed as the change of the image gradient along the tangent to obtain a scalar $q$ 
approximating $\kappa$. 
In addition,~\cite{Deng2007} presented PCBR, Principal Curvature-Based Regions, 
detector that uses the maximum or minimum eigenvalue of the Hessian matrix to find the principal curvature in a multi-scale space. 
As mentioned in~\cite{Deng2007}, the maximum eigenvalue yields a high value only for the dark side of edges, 
i.e. the minimum eigenvalue detects light lines on a dark background. 
By definition, it is a restrictive way to select features and it does not guarantee to select the maximal number of reliable features.

In conclusion of this review, curvature features have several advantages over more traditional intensity-based features~\cite{Fischer2014}, 
especially with extracting local structure of the points of interest. 
In addition, particularly, curvature features are invariant to viewpoint changes and to transformations that do not change the shape of the surface.

\subsection{Classic 3D feature detection}
In this subsection, our goal is to find if, in the literature, some 3D detector can be directly adapted to 2D, in order to obtain comparable points of interest.
Feature extraction of 3D models/scenes can be classified into point-based and image-based approaches.
Most of point-based methods are based on using SIFT in 3D by proposing an adaptation of the initial SIFT~\cite{Chang2005,Sattler2011}.
These approaches are interesting but they are not dedicated to 2D/3D registration and, so, they do not consider to detect similar features in 2D and in 3D.
Other methods proposed to find curves that have special properties in terms of differential geometry of the surface.
For example, in~\cite{Ohtake2004}, curves are supposed to be located at paraboloic lines which occur at points of extremal curvature.
These curves capture important object properties closed to the object surface, but they do not vanish along the surface when the viewpoint changes.

%
%
%

In image-based approaches, the 3D model is first rendered to form images or geometric buffers.
Then image processing methods are applied such as edge detection~\cite{Lee2007}.
In~\cite{Park2013}, 2D SIFT is applied on images of a rendered 3D mesh because this multi-scale representation extracts features 
that are supposed to be related to local extrema of the surface variation.
The apparent ridges (AR), which are a set of curves whose points are local maxima on a surface, are introduced in~\cite{Judd2007}.
In this paper, a view-dependent curvature corresponds to the variation of the surface normal with respect to a viewing screen plane.
However, apparent ridges often produce false edges that are not related to occluding contours, which are important for pose estimation of most of objects.
Mesh saliency measures the region importance of 3D models using Gaussian-weighted mean curvatures in multi-scales~\cite{Godil2011}.
However, mesh saliency is based on mesh images that are also affected by lighting conditions.
Average Shading Gradients, ASG, was proposed in~\cite{Plotz2015}. 
This rendering technique is based on averaging gradients over all lighting directions to cope with the unknown lighting conditions. 


In conclusion, only~\cite{Plotz2015} can be used in a 2D/3D registration context and we will describe, in section~\ref{sec:3DModelRepr}, and compare our work with this method, 
in section~\ref{exp}. 

\subsection{2D/3D registration and pose estimation}

%

In the computer vision literature, the problem of automatically aligning 2D photographs with an existing 3D model of the scene has been investigated in depth 
over the past fifteen years. 
In the general case, the proposed solution will be an image-to-model registration to estimate the 3D pose of the object. 
The 2D-to-3D registration problem is approached in the literature through indirect and direct methods~\cite{Paudel2014}.

For \textbf{indirect registration}, these methods are performed either by 3D-to-3D registration or by finding some appropriate registration parameters, 
such as the standard Iterative Closest Point, ICP algorithm~\cite{Besl1992}. 
This kind of techniques are more global and do not really provide points to points correspondences. 

For \textbf{direct registration methods}, in~\cite{Sattler2011}, correspondences are obtained by matching SIFT feature descriptors 
between SIFT points extracted in the images and in the 3D models. 
However, establishing reliable correspondences may be difficult due to the fact that the set of points in 2D and in 3D are not always similar, 
in particular because of the variability of the illumination conditions during the 2D and 3D acquisitions. 
In the same context, in~\cite{Yong2013}, the authors assume that the object in the input image has no or poor internal texture.
Methods relying on higher level features, such as lines~\cite{Xu2016}, planes~\cite{Tamaazousti2011} and building bounding boxes~\cite{Liu2005},
are generally suitable for Manhattan World scenes and hence applicable only in such environments.
Skyline-based methods~\cite{Ramalingam2009} as well as methods relying on a predefined 3D model~\cite{Clarkson2001} are, likewise, of limited applicability.
Recently, the histogram of gradients, HOG, detector~\cite{Aubry2014,Lim2014} or a fast version of HOG~\cite{Choy2015} have been also used to extract the features from rendering views and real images.
All of these approaches give interesting results, however, they do not evaluate the repeatability between the set of points detected in an intensity image and those detected in an image rendered from the 3D model.
Finally, in~\cite{Plotz2015}, 3D corner points are detected using the 3D Harris detector and the rendering average shading gradients images on each point.
For a query image, similarly, corner points are detected in multi-scale. 
Then, the gradients computed for patches around each point is matched with the database containing average shading gradient images using HOG descriptor.
This method still relies on extracting gradients of photographs affected by textures and background and it can give erroneous correspondences.
Consequently, they propose a refine stage based on RANdom SAmple Consensus, RANSAC~\cite{Fischler1981} to improve the final pose estimation.

In this paper, structural cues (e.g., curvilinear shapes) based on curvilinear saliency are extracted instead of only considering silhouettes,
since they are more robust to intensity, color, and pose variations.
In fact, they have the advantage to both represent outer and inner (self-occluding) contours that also characterize the object and that are useful for estimating the pose.
In order to merge in the same descriptor curvilinear saliency values and curvature orientation, the HOG descriptor, 
widely used in the literature and that properly describes the object shape, is employed.
Finally, HOG features and repeatability scores are used to match the query image with a set of depth images rendered from a 3D model.

In the rest of the paper, after the reminder on differential geometry, we will describe the 3D model representation and the image representation that are introduced in order to compare and to match the 3D and the 2D data. 
In particular, we will explain how these representations allow to be robust to background details and to texture before illustrating the interest of the proposition with 
experimental results on feature detection, registration between 2D images and 3D models and pose estimation. 

\section{Reminder on Differential Geometry}\label{sec:reminder}

\subsection*{Notations}

In the sequel, these notations are used: 
 \begin{itemize}
  \item $\bm\nabla_ f$: the gradient vector of a scalar-valued function $f$. 
  \item $\mathbf{F}_x$:  the partial first-order derivative $\frac{\partial\mathbf{F}}{\partial x}$ of a vector-valued
function $\mathbf{F}$ w.r.t. variable $x$. 
\item Similarly,  $\mathbf{F}_{xy}$: the partial second-order derivatives  $\frac{\partial^2\mathbf{F}}{\partial
x\partial y}$  of  $\mathbf{F}$ w.r.t. variables $x$ and $y$.
\end{itemize}

Moreover, we assume a calibrated
perspective camera where the image
point coordinates are given  with respect to the normalized image
 frame i.e., as $\mathbf{x} = (x,y)$ obtained from the equation
 $\left[x,y,1\right]=\mathtt{K}^{-1}\left[u,v,1\right]$, where $(u,v)$ are
 pixel coordinates and $\mathtt{K}$ is the usual calibration upper triangular matrix~\cite{Hartley2004}.

\subsection{Definitions: Differential of a Map -- Tangent Plane -- Gauss Map}
\label{ss:diff:map}
\begin{figure}[htb]
\begin{center} 
\begin{tabular}{c}
\!\!\!\!\!\!\includegraphics[width=1.05\columnwidth]{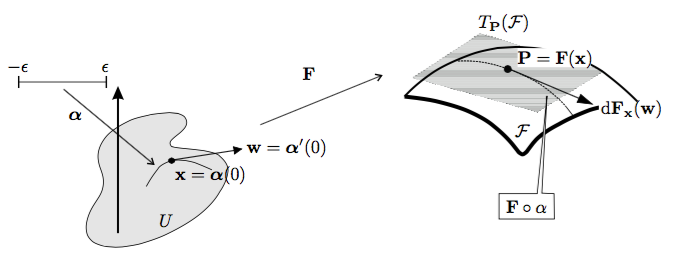}
 \\
(a)\\
\!\!\!\!\!\!\includegraphics[width=1.05\columnwidth]{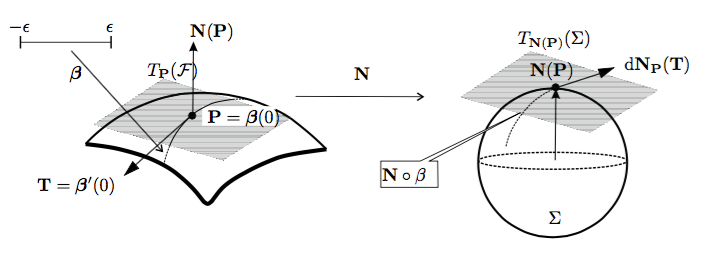}
 \\
(b)  
\end{tabular}
  \caption{\small (a) A map $\mathbf{F}$  parameterizing a regular surface  $\mathcal{F}$
 and its differential at point $\mathbf{x}$ along direction $\mathbf{w}$.
(b)  The map N is the Gauss map (which is a mapping between the  surface
 $\mathcal{F}$ and the unit sphere  $\Sigma$) and differential at  $\mathbf{P}$
along  $\mathbf{T}$.  }
  \label{fig:PG:EXF1}
\end{center}
\end{figure}

Let  $\mathcal{F}\subset \mathbb{R}^3$ be a
{\textit{regular surface}}\footnote{See the definition of a regular surface in $R^n$ in~\cite[p.~281-286]{Gray2006}. }
whose parameterization is given by the differentiable map
$\mathbf{F}:U\subset\mathbb{R}^2\rightarrow\mathcal{F}$ with 
\begin{equation}
\mathbf{F}(x,y)=[X(x,y),Y(x,y),Z(x,y)]^\top
\label{equ:09:13:01}
\end{equation}
To each
$\mathbf{x}=[x,y]^\top\in U$ is associated a  map   $\mathrm{d}{\mathbf{F}_\mathbf{x}}:\mathbb{R}^2\rightarrow\mathbb{R}^3$,
called the \textit{differential of $\mathbf{F}$ at $\mathbf{x}$} and  defined as follows~\cite[p.~128]{doCarmo2016}.
Let $\mathbf{v}\in \mathbb{R}^2$   be a vector and let 
$\bm\alpha:(-\epsilon,\epsilon)\rightarrow U$
  be a
differentiable  curve satisfying $\bm\alpha(0) = \mathbf{x}$ and $\bm\alpha'(0) = \mathbf{v}$.
By the chain rule, the  curve 
$\mathbf{F}\circ\bm\alpha$ 
in $\mathbb{R}^3$ is  also differentiable.  We then define\begin{equation*}
\mathrm{d}{\mathbf{F}_\mathbf{x}}( \mathbf{v})=(\mathbf{F}\circ\bm\alpha)'(0)
\end{equation*}
It provides a linear (i.e., first-order) approximation 
to $\mathbf{F}(\mathbf{x}+\mathbf{v})$
when the increment $\mathbf{v}$
is small enough.
This is illustrated in Fig.~\ref{fig:PG:EXF1}(a).
The vector subspace $\mathrm{d}{\mathbf{F}_\mathbf{x}}(U)\subset \mathbb{R}^3$
 has dimension 2 and is a plane consisting of all tangent vectors of $\mathbf{F}$
at $\mathbf{P}=\mathbf{F}(\mathbf{x})$.
It is called the \textit{tangent plane
 of $\mathcal{F}$ at $\mathbf{P}$ }and  denoted by
 $T_\mathbf{P}(\mathcal{F})$.

It can be proved~\cite[p129]{doCarmo2016} that the above
definition does not depend on the
choice of  $\bm\alpha$.
Furthermore, the fact that 
$(\mathbf{F}\circ\bm\alpha)'(0)=\mathbf{F}'(\bm\alpha(0))\bm\alpha'(0)$
entails that $\mathrm{d}{\mathbf{F}_\mathbf{x}}( \mathbf{v})$ is   linear
in $ \mathbf{v}$. In particular, in the  canonical bases of $\mathbb{R}^2$ and
$\mathbb{R}^3$, we have
$$
\mathrm{d}{\mathbf{F}_\mathbf{x}}(\mathbf{v})
=\mathtt{J}_\mathbf{F}(\mathbf{x})\mathbf{v}
$$
involving the $3\times 2$ \textit{Jacobian
 matrix 
of $\mathbf{F}$ at $\mathbf{x}$  } 
\begin{equation}\label{equ:123:010:1}
\mathtt{J}_\mathbf{F}(\mathbf{x})=
\begin{bmatrix}
\mathbf{F}_{x}(\mathbf{x}) & \mathbf{F}_{y}(\mathbf{x}) 
\end{bmatrix}
\end{equation}
 with  
$\mathbf{F}_{x}$ and $\mathbf{F}_{y}$ as columns. This also shows that the vector subspace $\mathrm{d}{\mathbf{F}_\mathbf{x}}(U)$
 has indeed dimension 2.

Let  $\mathbf{P}=\mathbf{F}(\mathbf{x})$ be a point of $\mathcal{F}$. Let $\mathbf{N}:\mathcal{F}\subset \mathbb{R}^3\rightarrow\Sigma\subset\mathbb{R}^3$ be
        the differentiable map that assigns  to 
$\mathbf{P}$ the
 coordinate vector  $\mathbf{N}(\mathbf{P})$ on the unit sphere $\Sigma$ representing
   the unit normal
of $\mathcal{F}$ at $\mathbf{P}$ and computed as
\begin{equation}\label{equ:PG:850}
\mathbf{N}(\mathbf{P})=\frac{
\mathbf{F}_x(\mathbf{x})\times\mathbf{F}_y(\mathbf{x})}
{\left\Vert
\mathbf{F}_x(\mathbf{x})\times\mathbf{F}_y(\mathbf{x})\right\Vert}
\text{ with }
\mathbf{x}=\mathbf{F}^{-1}(\mathbf{P})\end{equation}
    This map is called the  {\textit{Gauss map}} of  $\mathcal{F}$.

The Gauss map is a mapping between the two surfaces  
$\mathcal{F}$ and  $\Sigma$ and the definition of differential 
is extended to that case.
The \textit{differential
of the Gauss map of  $\mathcal{F}$ at point $\mathbf{P}$ }is the  map   
$\mathrm{d}{\mathbf{N}_\mathbf{P}}:T_\mathbf{P}(\mathcal{F})\subset\mathbb{R}^3\rightarrow\mathbb{R}^3$
  defined as follows.
Let $\mathbf{V}\in T_\mathbf{P}(\mathcal{F})$   be a vector on the tangent plane
 of $\mathcal{F}$ at $\mathbf{P}$ and let 
$\bm\beta :(-\epsilon,\epsilon)\rightarrow\mathcal{F}$
  be a
differentiable  curve  on the surface  $\mathcal{F}$ satisfying $\bm\beta(0) = \mathbf{P}$ and $\bm\beta'(0)
= \mathbf{V}$.
By the chain rule, the  curve 
$\mathbf{N}\circ\bm\beta$ 
in $\mathbb{R}^3$ is  also differentiable;  we then define\begin{equation*}
\mathrm{d}{\mathbf{N}_\mathbf{P}}( \mathbf{V})=(\mathbf{N}\circ\bm\beta)'(0)
\end{equation*}
It expresses how $\mathbf{N}$  behaves
 --- how  $\mathcal{F}$ curves--- in the vinicity of $\mathbf{P}$.
This is illustrated in Fig.~\ref{fig:PG:EXF1}(b).
Again, it can be proved~\cite[p129]{doCarmo2016} that the above
definition of  $\mathrm{d}{\mathbf{N}_\mathbf{P}}$
does not depend on the
choice of one possible curve   $\bm\beta$.
 
Similar to the differential of a map seen above, the fact that 
$(\mathbf{N}\circ\bm\beta)'(0)=\mathbf{N}'(\bm\beta(0))\bm\beta'(0)$
entails that $\mathrm{d}{\mathbf{N}_\mathbf{P}}( \mathbf{V})$ is   linear
in $ \mathbf{V}=\bm\beta'(0)$. The vector subspace $\mathrm{d}{\mathbf{N}_\mathbf{P}}(T_\mathbf{P}(\mathcal{F}))\subset \mathbb{R}^3$
 has dimension 2: it is the plane  
 $T_\mathbf{N(P)}(\Sigma)$ consisting of all tangent vectors  to the unit
 sphere at point $\mathbf{N}(\mathbf{P})$ called the tangent plane of $\Sigma$ at $\mathbf{N}(\mathbf{P})$.
Therefore the domain of values of $\mathrm{d}{\mathbf{N}_\mathbf{P}}$ is $\mathrm{d}{\mathbf{N}_\mathbf{P}}(T_\mathbf{P}(\mathcal{F}))=T_\mathbf{N(P)}(\Sigma)$. Actually, it can readily seen that $T_\mathbf{N(P)}(\Sigma)$ and $T_\mathbf{P}(\mathcal{F})$ are parallel planes so the differential   of 
$\mathbf{N}$ is usually defined as  $\mathrm{d}{\mathbf{N}_\mathbf{P}}:T_\mathbf{P}(\mathcal{F})\rightarrow
T_\mathbf{P}(\mathcal{F})$.

\subsection{Curvatures. Fundamental forms of a surface}
Let $\mathbf{T}\in T_\mathbf{P}(\mathcal{F})$   be a \textit{unit vector} representing a direction on the tangent
plane
 of $\mathcal{F}$ at  $\mathbf{P}$.
Let
 $\mathscr{C}$  be the  curve obtained by slicing
$\mathcal{F}$
 with the  normal section of 
$\mathcal{F}$
   at $\mathbf{P}$ along 
$\mathbf{T}$   i.e.,\footnote{This holds for any plane through $\mathbf{P}$
 parallel
to    $\mathbf{T}$.
 This is due to
 Meusnier's theorem~\cite[p482]{Gallier2001}  ``All curves lying on a surface
$\mathcal{S}$
 and having at a given point $\mathbf{P}\in \mathcal{S}$ the
 same tangent line have at this point the
 same normal curvatures. ''}  the plane through $\mathbf{P}$  parallel to
both
$\mathbf{N(P)}$ and   $\mathbf{T}$.
 The \textit{normal curvature}  of $\mathcal{F}$ in   (unit) direction
$\mathbf{T}$
is  the curvature of  $\mathscr{C}$  at  $\mathbf{P}$   which can be given
by~\cite[p. 144]{doCarmo2016}: 
\begin{equation}\label{EQU:PG:191}
\kappa_{N}(\mathbf{T})
=-\mathbf{T}\cdot\mathrm{d}{\mathbf{N}_\mathbf{P}}(\mathbf{T})
\end{equation}
It expresses how $\mathbf{N}$  behaves
 --- how  $\mathcal{F}$ curves--- in the vinicity of $\mathbf{P}$.
An important remark is that the \textit{radius of curvature} of  $\mathscr{C}$   at
 $\mathbf{P}$ is  equal to 
$
1/\vert\kappa_N\vert
$.

The \textit{{principal curvatures}}  $\kappa_1$, $\kappa_2$ of $\mathcal{F}$
  at $\mathbf{P}$ can
be defined as the extrema of  function (\ref{EQU:PG:191}) 
with respect to directions $\mathbf{T}\in T_\mathbf{P}(\mathcal{F})$, subject to the constraint $\left\Vert \mathbf{T}\right\Vert=1$.
 The corresponding
directions are called
{\textit{principal directions}}  of $\mathcal{F}$ at $\mathbf{P}$.
It is well-known that `\textit{`the coefficients $\kappa_1$, $\kappa_2$ are  decisive
parameters that fully
describe local surface shape up to the second order
modulo a rigid movement.}''~\cite{Koenderink1992}
This means that the principal curvatures are invariant to the surface parameterization.

Now consider a new 3D coordinate system with
$\mathbf{P}=\mathbf{F}(\mathbf{x})$ as  origin.
Any vector $\mathbf{T}$ on  the tangent plane $T_\mathbf{P}(\mathcal{F})$ can
be written as
\begin{equation}
\mathbf{T}=u\mathbf{F}_x(\mathbf{x})+v\mathbf{F}_y(\mathbf{x})=\mathtt{J}_\mathbf{F}(\mathbf{x})\begin{bmatrix}u
\\
v \\
\end{bmatrix}\label{EQU:PG:133}
\end{equation}
where $\mathtt{J}_\mathbf{F}(\mathbf{x})$ is the Jacobian matrix of $\mathbf{F}$ defined in (\ref{equ:123:010:1})à and $(u,v)$ are  so-called \textit{local coordinates} of  $\mathbf{T}$.
From now on, we will put $uv$ as subscript to relate a vector to its local
coordinates, e.g.,   $\mathbf{T}_{uv}$.  

\subsubsection{First fundamental form of a surface}
Given
any $(u,v)$, the norm of any vector  $\mathbf{T}_{uv}$ on  $T_\mathbf{P}(\mathcal{F})$
writes
\begin{equation}\label{EQU:PG:180}
\left\Vert \mathbf{T}_{uv}\right\Vert=\sqrt{[u,v]\mathtt{I}_\mathbf{P}[u,v]^\top}
\end{equation}
where
\begin{align}\label{equ:PG:901}
\mathtt{I}_\mathbf{P}
&=
\begin{bmatrix}
\mathbf{F}_x(\mathbf{x})\cdot\mathbf{F}_x(\mathbf{x}) & \mathbf{F}_x(\mathbf{x})\cdot\mathbf{F}_y(\mathbf{x})
\\
\mathbf{F}_x(\mathbf{x})\cdot\mathbf{F}_y(\mathbf{x}) & \mathbf{F}_y(\mathbf{x})\cdot\mathbf{F}_y(\mathbf{x})
\end{bmatrix}
\notag
\end{align}
The quadratic form on  $T_\mathbf{P}(\mathcal{F})$ 
\begin{equation}\label{EQU:PG:182}
\mathfrak{I}_\mathbf{P}(u,v)\doteq{[u,v]\mathtt{I}_\mathbf{p}[u,v]^\top}
\end{equation}
%
 is called the   \textit{{first fundamental form} of $\mathcal{F}$}
\cite[p94]{doCarmo2016}.

\subsubsection{Second fundamental form of a surface}
 Let  $\mathbf{T}_{uv}$ be a direction of  the tangent plane $T_\mathbf{P}(\mathcal{F})$, given in local 2D coordinates.  On the one hand, it can be shown~\cite[p156]{doCarmo2016} that the differential
of the Gauss map of  $\mathcal{F}$ at $\mathbf{P}=\mathbf{F}(\mathbf{x})$ along   $\mathbf{T}_{uv}$
writes in standard 3D coordinates
$$\mathrm{d}{\mathbf{N}_\mathbf{P}}(\mathbf{T}_{uv})=\left[\mathbf{N}_{x}(\mathbf{x})\mid\mathbf{N}_{y}(\mathbf{x})\right]\begin{bmatrix}u \\
v \\
\end{bmatrix}$$
On the other hand,
let denote by $\mathrm{d}{\mathbf{N}_\mathbf{P}}(u,v)$ 
the differential  of 
$\mathbf{N}$ at $\mathbf{P}=\mathbf{F}(\mathbf{x})$ along   $\mathbf{T}_{uv}$ {expressed in local 2D coordinates}
i.e., such that $\mathrm{d}{\mathbf{N}_\mathbf{P}}(\mathbf{T}_{uv})
=\mathtt{J}_\mathbf{F}(\mathbf{x})]\mathrm{d}{\mathbf{N}_\mathbf{P}}(u,v)$.
Then we have
\begin{equation}
\mathrm{d}{\mathbf{N}_\mathbf{P}}(u,v)=\mathtt{I}_\mathbf{P}^{-1}\mathtt{II}_\mathbf{P}
\begin{bmatrix}u \\
v \\
\end{bmatrix}
\end{equation}
where
\begin{align}
\mathtt{II}_\mathbf{P}
&=
\begin{bmatrix}
\mathbf{N}_x(\mathbf{x})\cdot \mathbf{F}_{x}(\mathbf{x}) & \mathbf{N}_x(\mathbf{x})\cdot \mathbf{F}_{y}(\mathbf{x})\\
\mathbf{N}_x(\mathbf{x})\cdot \mathbf{F}_{y}(\mathbf{x}) & \mathbf{N}_y(\mathbf{x})\cdot \mathbf{F}_{y}(\mathbf{x})
\end{bmatrix}
\end{align}
The proof can be found in~\cite[p156]{doCarmo2016}.
%

The quadratic form
\begin{equation}\label{EQU:PG:182bis}
\mathfrak{I\!I}_\mathbf{P}(u,v)={[u,v]\mathtt{II}_\mathbf{P}[u,v]^\top} 
\end{equation}
%
 is called the   \textit{{second fundamental form} of $\mathcal{F}$}
\cite[p143]{doCarmo2016}.
It directly follows from this that Eq.\ (\ref{EQU:PG:191})  can be expressed in local coordinates and writes
\begin{equation}\label{EQU:PG:192}
\kappa_{N}(u,v)={[u,v]\mathtt{II}_\mathbf{P}[u,v]^\top} \end{equation}
For any (non unit)  vector $\mathbf{V}(u,v)$ on
 $T_\mathbf{P}(\mathcal{S})$, in order to set it as a unit direction, it is needed to divide (\ref{EQU:PG:192})
by the square of expression  (\ref{EQU:PG:180}). Hence, the \textit{normal curvature}
in  the unit direction $\frac{\mathbf{V}}{\left\Vert \mathbf{V}\right\Vert}$
 is can be given by
\begin{equation}\label{equ:PG:480}
\kappa_{N}(u,v)=-\frac{\mathbf{V}_{uv}\cdot\mathrm{d}{\mathbf{N}_\mathbf{P}}(\mathbf{V}_{uv})}{\left\Vert
\mathbf{V}_{uv}\right\Vert^2}=\frac{\mathfrak{I\!I}_\mathbf{P}(u,v)}{\mathfrak{I}_\mathbf{P}(u,v)}
\end{equation}

 \subsubsection{Closed-form solutions for principal curvatures}
The {{principal curvatures}}  $\kappa_1$, $\kappa_2$ of $\mathcal{S}$   at $\mathbf{P}$ can
be defined as the extrema of  function (\ref{equ:PG:480}) with  $(u,v)$-coordinates
as variables.

Seeing (\ref{equ:PG:480}) as a generalized Rayleigh quotient, it
is known~\cite[p18]{Parlett1998} that $\kappa_{{N}}$ has an extremum
at $(\hat u,\hat v)$ only if $\kappa_{{N}}(\hat u,\hat v)$
is a root of $\det(^{}\mathtt{II}_\mathbf{P}-\kappa_{{N}}\mathtt{I}_\mathbf{P})$
 or, equivalently, only if $\kappa_{{N}}(\hat u,\hat v)$ is an eigenvalue
of the $2\times 2 $ matrix  $\mathtt{I}_\mathbf{P}^{-1}\mathtt{II}_\mathbf{P}$, which is \textit{not}
symmetric but always
 has real eigenvalues~\cite[p500]{Gallier2001}. 
As a result, the principal curvatures are the two eigenvalues  $\kappa_\alpha$   ($\alpha=1,2$) of the matrix $\mathtt{I}_\mathbf{P}^{-1}\mathtt{II}_\mathbf{P}$. The principal 3D directions are  
$\mathbf{T}_\alpha=\mathtt{J}_\mathbf{F}(\mathbf{x})\mathbf{e}_\alpha$ where  $\mathbf{e}_\alpha$  are  the corresponding eigenvectors. 

Now we state a proposition that we derive from  the above results, which will be used in our work.

\begin{proposition}\label{prop:000:123}
The principal curvature $\kappa_\alpha$   ($\alpha=1,2$) at $\mathbf{P}$
 associated to  the unit principal 3D direction   
$\mathbf{T}_\alpha$
 is equal to the absolute magnitude
of the differential of the  Gaussian
map at this point   i.e.,
\begin{equation}\label{EQUPG:00}
\left\vert\kappa_\alpha\right\vert
=\left\Vert \mathrm{d}{\mathbf{N}_\mathbf{P}}(\mathbf{T}_\alpha)\right\Vert
\end{equation}

\end{proposition}
\begin{myproof}
Since the Euclidean norm is invariant to changes of Euclidean coordinates, without loss of generality, let choose a new parameterization
$\mathbf{\tilde S}(\tilde x,\tilde y)=[ \tilde x,\tilde y,{\tilde Z}(\tilde
x,\tilde
y) ]^\top$, for some  new height function $\tilde Z$, w.r.t. 3D orthonormal  frame whose origin is   $\mathbf{P}$ and  $\tilde x\tilde y$-plane
coincides with
the tangent plane $T_\mathbf{P}(\mathcal{F})$.
%
%
More generally,  we will add the symbol $\,\widetilde{}\,$ to distinguish the new representations from the old ones, except for the principal curvatures which are irrespective of  parameterizations. Let us remind that 
$\mathbf{\tilde T}_\alpha=\mathtt{J}_\mathbf{F}(\mathbf{x})\mathbf{\tilde  e}_\alpha$,
where $\mathbf{\tilde  e}_\alpha$ is the associated eigenvector, and note that the new    first fundamental matrix,
    $\mathtt{\tilde I}_\mathbf{P}$, is then the identity. As a result, starting from the fact that  $\kappa_\alpha$ is an eigenvalue
of the $2\times 2 $ matrix  $\widetilde{\mathtt{I}}_\mathbf{\tilde P}^{-1}\widetilde{\mathtt{II}}_\mathbf{\tilde
P}$, we have:
\begin{align*}
(\widetilde{\mathtt{I}}_\mathbf{\tilde P}^{-1}\widetilde{\mathtt{II}}_\mathbf{\tilde P})\mathbf{\tilde e}_\alpha
&=\kappa_\alpha \mathbf{\tilde e}_\alpha\\
\Leftrightarrow\mathrm{d}{\mathbf{\tilde N}_\mathbf{\tilde P}}
(\mathbf{\tilde  e}_\alpha)
&=\kappa_\alpha\mathbf{\tilde e}_\alpha\\
\Leftrightarrow\mathrm{[\mathbf{\tilde S}_{\tilde x}\mid\mathbf{\tilde
S}_{\tilde y}]d}{\mathbf{\tilde N}_\mathbf{\tilde P}}(\mathbf{\tilde  e}_\alpha)
&=\kappa_\alpha[\mathbf{\tilde S}_{\tilde x}\mid\mathbf{\tilde
S}_{\tilde y}]\mathbf{\tilde e}\\
\Leftrightarrow \mathrm{d}{\mathbf{\tilde N}_\mathbf{P}}(\mathbf{\tilde T}_\alpha)&=\kappa_\alpha\mathbf{\tilde T}_\alpha\\
\Rightarrow \Vert \mathrm{d}{\mathbf{\tilde N}_\mathbf{P}}(\mathbf{\tilde T}_\alpha)\Vert^2
&=\kappa_\alpha^2
\end{align*}
\end{myproof}

\section{3D model representation}\label{sec:3DModelRepr}

The work the most related to what is proposed in this paper is the Average Shading Gradient (ASG) approach,  proposed in~\cite{Plotz2015}. 
After introducing how object surface can be represented, we highlight the differences between these two approaches. 

\noindent\textbf{Object surface : }
Denote by   $\mathcal{M}$   the surface of some observed object associated to a parameterization  $\mathbf{M}(x,y)\triangleq[X(x,y),Y(x,y),Z(x,y)]^\top$,
  where $(x,y)$ varies over
the restricted image domain of a given camera which is delimited by the occluding contour  of
 the object.
Under perspective projection, every visible 3D point of  $\mathcal{M}$
(seen from the camera viewpoint), with vector
$\mathbf{M}(x,y)$, is assumed to be in one-to-one correspondence
with the 2D image point with vector $\mathbf{x}=[x,y]^\top$,
such that $x=X(x,y)/Z(x,y)$ and $y=Y(x,y)/Z(x,y)$.
As a result, we get
 \begin{equation}
\mathbf{M}(x,y)=Z(x,y)[x,y,1]^\top\label{equ:PG:001}
\end{equation}
Let $\mathbf{N}(x,y)$ denotes the  \textit{Gaussian map} of $\mathcal{M}$  which  assigns, on the
unit sphere, to each point point
$\mathbf{M}(x,y)$ of  $\mathcal{M}$ the unit normal
of $\mathcal{M}$ defined by 
$\mathbf{N}(x,y)=\frac{\mathbf{\bar
N}(x,y)}{\left\Vert\mathbf{\bar N}(x,y)\right\Vert}$ where, using~(\ref{equ:PG:001}),
 $\mathbf{\bar N}$ writes
\begin{equation}\label{equ:PG:300}
\mathbf{\bar N}
=
\mathbf{M}_x\times\mathbf{M}_y \\
=Z\left[ -Z_x, -Z_y, xZ_x+yZ_y+Z \right]^\top
\end{equation}
It can be shown that the  Jacobian $3\times 2$ matrix of $\mathbf{N}$ writes
\begin{equation}\label{EQU:432:001}
\mathtt{J}_\mathbf{N}=
\begin{bmatrix}
\mathbf{N}_x &
\mathbf{N}_y
\end{bmatrix}
=\left( \mathtt{I}-{\mathbf{N}\mathbf{N}^\top}\right)\mathtt{J}_\mathbf{\bar N}
\end{equation}
where  the columns of 
$\mathtt{J}_\mathbf{\bar N}=
\begin{bmatrix}
\mathbf{\bar N}_x &
\mathbf{\bar N}_y
\end{bmatrix}
$
have the form
\begin{equation}
\label{EQU:432:002}
\mathbf{\bar N}_\star=\begin{bmatrix}
Z_{x}Z_{\star}-Z_{x\star}Z  \\
Z_{\star}Z_{y}-Z_{\star y}Z  \\
 xZ_{x\star}Z+yZ_{\star y}Z+Z_{\star}(xZ_x+yZ_y+3Z)  \\
\end{bmatrix} 
\end{equation}
$\star$ standing for either $x$ and or $y$.


%
%
%
%
%
%
%

\subsection{Average Shading Gradient (ASG) Feature~\cite{Plotz2015}}
Pl\"otz \textit{et al.} assumed in~\cite{Plotz2015}
that the image intensity function
obeys the Lambertian shading function
\begin{equation}\label{equ:PG:1001}
I(x,y)\propto\max(0,-\mathbf{N}(x,y)\cdot\mathbf{s})
\end{equation}
for a parallel light source $\mathbf{s}\in\mathbb{R}^3$.
This means that the reflectance  describing the object material is assumed
 to be Lambertian with constant
albedo\footnote{A general shading function is
$I(x,y)=\rho(\mathbf{M}(x,y))\max(0,-\mathbf{N}(x,y)\cdot\mathbf{s})$
where $\rho(\mathbf{M}(x,y))$ is the albedo at object  point $\mathbf{M}(x,y)$.
}.
In addition, the background is assumed to be constant (e.g., a plane).

The authors propose as feature in the intensity image the magnitude of the gradient of the shading function.  
To register the intensity image to the 3D (untextured) model,  the idea is to generate virtual images viewing the object from different camera pose candidates. 
Nevertheless, it is clearly impossible to render any such virtual image obeying the shading function~(\ref{equ:PG:1001}) 
without prior information about the lighting direction and so about $\mathbf{s}$.
Therefore, the authors propose to  replace the gradient magnitude feature, in the virtual images, 
by a feature corresponding to the average value of the gradient magnitude computed over all light directions, so-called \textit{average shading gradient} magnitude.  
Denoting~$\left\Vert\bm\nabla_ I\right\Vert^2 = {I_x^2+I_y^2}$ the magnitude of the gradient of the shading function~(\ref{equ:PG:1001}) 
then the magnitude of the average shading gradient is:
\begin{equation}\label{eq:image}
\overline{\left\Vert\bm\nabla_ I\right\Vert}=
\int_{\mathcal{S}} \left\Vert\bm\nabla_I\right\Vert \mathrm{d}\mathbf{s}
\\
\end{equation}
where the  vector $\mathbf{s}$, cf.\ (\ref{equ:PG:1001}),
varies over the unit sphere ${\mathcal{S}}$ in $\mathbb{R}^3$ and $\mathrm{d}\mathbf{s}$
is the volume element.

The nice contribution of Pl\"otz \textit{et al.}  is, by applying Jensen's
inequality, to derive the following closed-form bound on $\overline{\left\Vert\bm\nabla_
I\right\Vert}$
\begin{align}
\overline{\left\Vert\bm\nabla_ I\right\Vert}
&\le
\sqrt{\int_{\mathcal{S}} \left\Vert\bm\nabla_I\right\Vert^2 \mathrm{d}\mathbf{s}
}\notag\\
&=\gamma\sqrt{
\left(\left\Vert\mathbf{N}_x\right\Vert^2+\left\Vert\mathbf{N}_y\right\Vert^2\right)
}\label{equ:PG:103}
\end{align}
with $\gamma=\sqrt{\frac{\pi}{3}}$.
It is reported by the authors to behave like a very good approximation of
$\overline{\left\Vert\bm\nabla_ I\right\Vert}$. This is the elegant way
 the authors get rid of the unknown lighting direction
 $\mathbf{s}$ in (\ref{equ:PG:1001}).



\subsection{Curvilinear Saliency Features (CS)}
\label{prop:curv:feat}
As already mentioned, our goal is to find a common representation between the 3D model and the 2D image in order to be able to match them. 
For that purpose, we first show how the 3D model can be represented or studied from different points of view and 
how these different viewpoints can be analyzed and compared to a 2D image. 
For that purpose, we   represent the observed 3D object  by a set of  synthetic
depth maps generated from  camera locations distributed on concentric spheres encapsulating, by sampling elevation and azimuth angles, as well
as distances from the camera to the object. 
A depth map (or  depth image)  $Z(x,y)$ associates
to every image point $(x,y)$ the $Z$-coordinate, w.r.t. the camera frame,
of
the object 3D point (\ref{equ:PG:001}) that projects onto $(x,y)$.

Let $\mathcal{D}$   denote the depth surface  that is the 3D surface whose
graph parameterization is\footnote{Note the difference with (\ref{equ:PG:001}).
}
 $$\mathbf{D}(x,y)=\left[ x,y,Z(x,y) \right]^\top$$

Which features should be extracted in the depth map? We aim at detecting depth
``discontinuities''  by searching points on $\mathcal{D}$  having high principal
curvature in one direction
and low principal curvature in the orthogonal direction.  
We call \textit{Curvilinear Saliency features} of a surface
  loci of such points. Basically, they correspond to the \textit{ridges}
and \textit{valleys} of this surface.
In this work, we use the difference of the principal
curvatures $\kappa_1 - \kappa_2$ to describe the ridges and
valleys.



\noindent\textbf{Principal curvatures and directions: }
Consider  a point $\mathbf{P}=\mathbf{D}(x,y)$.
 Let $\mathbf{N'}(x,y)$ denote the  \textit{Gaussian map} 
 of $\mathcal{D}$
assigning to $\mathbf{P}$ the unit normal of  $\mathcal{D}$ at $\mathbf{P}$,
such that
\begin{equation}\label{equ:PG:440}
\mathbf{N'}=\frac{\mathbf{\bar N'}}
{\left\Vert\mathbf{\bar N'}\right\Vert}
\text{ where }
\mathbf{\bar N'}
=
\mathbf{D}_x\times\mathbf{D}_y
=\alpha
\begin{bmatrix}
-\bm\nabla_Z \\
1 \\
\end{bmatrix}
\end{equation}
with $\bm\nabla_ Z=[Z_x,Z_y]^\top$ and $\alpha =1/\sqrt{1+\left\Vert\bm\nabla_ Z\right\Vert^2}$.

 As the two columns of the Jacobian matrix
$\mathtt{J}_\mathbf{D}$ of $\mathcal{D}$
are 
   $\mathbf{D}_x=[1,0,Z_x]^\top$
and $\mathbf{D}_y=[0,1,Z_y]^\top$, the \textit{first fundamental form of $\mathcal{D}$}
can be computed as
\begin{equation}\label{equ:PG:901}
\mathtt{I}_\mathbf{P}
=\mathsf{I}_3+\bm\nabla_ Z\bm\nabla_ Z^\top
\notag
\end{equation}
and the   \textit{second fundamental form of $\mathcal{D}$}
can be computed as
%
\begin{equation}
\mathtt{II}_\mathbf{P}
=\alpha
\mathtt{H}_Z
\end{equation}
where $\mathtt{H}_Z$
is the Hessian matrix of $Z$ i.e., with the second-order partial derivatives of $Z$ w.r.t. $x$ and $y$ as
elements.


The \textit{principal curvatures} 
of  $\mathcal{D}$      at $\mathbf{P}$ coincide with  the eigenvalues
$\kappa_\alpha$ ($\alpha=1,2$) of $\mathtt{I}_\mathbf{P}^{-1}\mathtt{II}_\mathbf{P}$, which are always real. In the tangent plane $T_\mathbf{P}(\mathcal{D})$, 
the local coordinates of the  \textit{principal directions} of $\mathcal{D}$ at $\mathbf{P}$ are given by the  eigenvectors 
$\mathbf{e}_\alpha$ of $\mathtt{I}_\mathbf{p}^{-1}\mathtt{II}_\mathbf{p}$ so the 3D principal directions in 3D wrote $\mathtt{J}_\mathbf{D}\mathbf{e}_\alpha$. 
As Koenderink wrotes in~\cite{Koenderink1992}, 
``\textit{it is perhaps not superfluous to remark here that the simple (eigen-)interpretation in terms\footnote{By neglecting $\mathtt{I}_\mathbf{P}$.} 
of $\mathtt{II}_\mathbf{P}=\alpha\mathtt{H}_Z$ is only valid in representations where $\bm\nabla_Z=\mathbf{0}$}'', 
which is the condition for the point to be local extremum.

Thanks to proposition~\ref{prop:000:123}, we know that that the principal curvature $\kappa_\alpha$   at $\mathbf{P}$  associated to the principal 3D direction   
$\mathbf{T}_\alpha=\mathtt{J}_\mathbf{D}\mathbf{e}_\alpha$ is equal to the absolute magnitude of the change of the  normal
\begin{equation}\label{EQUPG:00bis}
\left\vert\kappa_\alpha\right\vert=\left\Vert \mathrm{d}{\mathbf{N}'_\mathbf{P}}(\mathbf{T}_\alpha) \right\Vert
\end{equation}
where  $\mathrm{d}{\mathbf{N}'_\mathbf{P}}(\mathbf{T})$ denotes the  differential  of  $\mathbf{N}'$ at $\mathbf{P}$ in  direction  $\mathbf{T}$.
We will make us of this result for the image representation, cf.~\S\ref{sec:ima:rep}. 
Now let us explain why we propose as feature the difference $\kappa_1-\kappa_2$ where $\kappa_1\ge\kappa_2$.

\noindent\textbf{Curvilinear feature:}
Without losing generality, let $\kappa_1$ and $\kappa_2$ be the  principal curvatures computed as ordered eigenvalues of $\mathtt{I}_\mathbf{p}^{-1}\mathtt{II}_\mathbf{p}$
so that $\kappa_1\ge\kappa_2$. 
We aim at detecting  points lying on ``elongated'' surface parts.
In this work, we detect points at which this difference is  high: 
\begin{equation}\label{equ:PG:101}
CS(x,y)={\kappa_1(x,y)-\kappa_2(x,y)}
\end{equation}
We call (\ref{equ:PG:101}) the \textit{curvilinear saliency (CS)} feature. Curvilinear means a feature that belongs to a curved line. 
The rest of this paragraph justifies such a choice.

\begin{figure}[h]
  \centering
  \includegraphics[width=0.7\columnwidth]{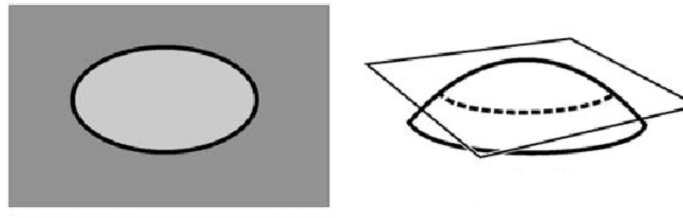}\\
  \includegraphics[width=0.7\columnwidth]{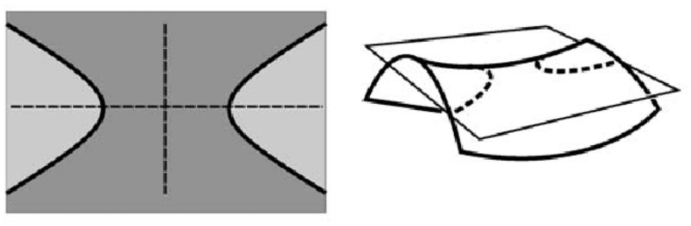}\\
  \includegraphics[width=0.7\columnwidth]{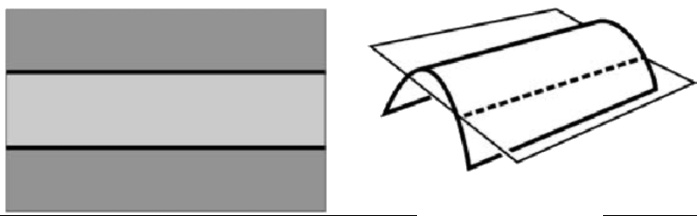}\\
  \caption{The real conics of the Dupin indicatrix.}\label{fig:PG:EX1}
  \label{tensor}
\end{figure}

Given a point $\mathbf{P}$ on $\mathcal{D}$, let $(\tilde x,\tilde y)$ be the Cartesian coordinates on the tangent plane $T_\mathbf{P}(\mathcal{D})$)
w.r.t. the 2D  frame whose origin is $\mathbf{P}$ and the orthonormal basis is formed by the principal directions $\{\mathbf{e}_1,\mathbf{e}_2\}$. 
As a result, $\mathcal{D}$ can now locally be associated to the new parameterization $\mathbf{F}(\tilde x,\tilde y)=\left[ \tilde x,\tilde y,F(\tilde x,\tilde y) \right]^\top$, 
for some  height function $F$. 
In that case, it can be readily seen that $\mathtt{I}_\mathbf{P}$ is the identity matrix, and so 
$\mathtt{I}_\mathbf{P}^{-1}\mathtt{II}_\mathbf{P}=\mathtt{II}_\mathbf{P}=\Diag(\kappa_1,\kappa_2)$ is exactly the Hessian matrix of $F$.
For some $\epsilon>0$  small enough, consider on  the two planes  parallel to $T_\mathbf{P}(\mathcal{D})$ at distances  $\pm\epsilon$ from $T_\mathbf{P}(\mathcal{D})$, 
the curves $\mathscr{C}_\pm=\{ (\tilde x,\tilde y),  \mathbf{F}(\tilde x,\tilde y)\in T_\mathbf{P}(\mathcal{D}) \mid F(\tilde x,\tilde y)=\pm\epsilon\}$. 
It can be shown~\cite[p500]{Gallier2001} that   the first-order approximation of the  intersections of $\mathcal{D}$  with the two parallel planes is 
the union of two conics  (one real and one virtual) with equations \(\mathtt{II}_\mathbf{P}(\tilde x,\tilde y)=\pm2\epsilon.\)
This union is known as the \textit{Dupin indicatrix} when written in canonical form (i.e., by replacing $2\epsilon$ by $1$).
%
The real Dupin conic characterizes the local shape of $\mathcal{D}$ and gives local information on the first-order geometry of the surface, 
at least at  points where the conic is non degenerate. It specializes to a parabola if the Gauss curvature vanishes i.e., $\kappa_1\kappa_2=0$, 
to an ellipse if $\kappa_1\kappa_2>0$, and  to an hyperbola if $\kappa_1\kappa_2<0$, see Fig.~\ref{fig:PG:EX1}. 
Points are said to be elliptic, hyperbolic or parabolic respectively.

Let us now focus on the  Dupin central conics i.e., the real ellipse and real hyperbola.
We do not consider the case of a parabola since, in real experiments, the condition $\kappa_2=0$ will never be verified exactly.

Various measures can describe such a  conic shape~\cite{Koenderink1992}.
We introduce the $CS$ quantity~(\ref{equ:PG:101})  that provides a  unified way of treating ellipses and  hyperbolas using the following nice interpretation.

Let the semi-major and semi-minor axes of the Dupin  central conic be $r_{\text{maj}}=\max(\rho_1 ^{2}, \rho_2 ^{2})$ 
and $r_\text{min}=\min(\rho_1 ^{2}, \rho_2 ^{2})$ respectively, where   $\rho_1$ and $\rho_2$ are the radii of curvature of the curves obtained 
through the  normal section of $\mathcal{D}$ at $\mathbf{P}$ along the principal directions.

\begin{proposition}
$CS$ in~(\ref{equ:PG:101})  is the squared ratio between the eccentricity $e$ of the Dupin
 conic and  its minor axis (due to lack of space, the straightforward proof is omitted): 

\begin{equation}\label{equ:PG:110}
\kappa_1-\kappa_2=\left(\frac{e}{r_{\text{min}}}\right)^2
\end{equation}
where
\begin{equation*}\label{equ:PG:110a}
e=\sqrt{
1 \pm\ \frac{r_{\text{min}}^2}{r_{\text{maj}}^2}}
\;\text{with}
\begin{cases}
-: \text {if the conic is an ellipse}\\
+: \text {if the conic is an hyperbola}
\end{cases}
\end{equation*} 
 
\end{proposition}

The eccentricity can  be interpreted as the fraction of the distance along the semimajor axis at which the focus lies.
%
The $CS $ quantity is normalized by additionally dividing the squared eccentricity  by the squared semi-minor axis.
Note that (\ref{equ:PG:110}) works for Dupin ellipses as well as Dupin hyperbolas.
The  curvilinear saliency $CS$  is large when $\kappa_1 \gg \kappa_2$, that is in presence of distant foci and so a highly elongated ellipse or a ``squashed''
hyperbola, see Fig.~\ref{fig:PG:EX1}.  
This occurs e.g., when the point is located on a depth ``discontinuity''.
In turn, when $\kappa_1 \simeq \kappa_2$, the conic approaches a circle and the distance between foci becomes very small.

\noindent\textbf{A simple way to compute the curvilinear  feature: }  
After  algebraic manipulations, it can be  shown   that
$\mathtt{I}_\mathbf{P}^{-1}\mathtt{II}_\mathbf{P}=\frac{1}{\alpha}\mathsf{M}$
where
\begin{equation}\label{equ:PG:007}
\mathsf{M}\triangleq\begin{bmatrix}
(Z^2_y  + 1) Z_{xx} - Z_x Z_y Z_{xy} & (Z^2_y  + 1)Z_{xy} - Z_x Z_y Z_{yy}
\\
(Z^2_x  + 1) Z_{xy} - Z_x Z_y Z_{xx} & (Z^2_x  + 1)Z_{yy} - Z_x Z_y Z_{xy}
\end{bmatrix}
\end{equation}

\begin{proposition}
 The squared curvilinear feature can be  computed as
\begin{align}
\label{lambdasD}
    \operatorname*{CS^2} &\triangleq\left\Vert \bm \nabla_Z \right\Vert^2\left(
(\Tr\mathsf{M})^2-4\det\mathsf{M} \right)\\
                         &=4\left\Vert \bm \nabla_Z \right\Vert^2(\bar{\kappa}^2-K)
\end{align}
where $\bar{\kappa}$ is the mean curvature of  $\mathcal{D}$ and $K$ its
Gaussian curvature.
\end{proposition}
\begin{myproof}
Let
$\lambda_1$ and $\lambda_2$ be the ordered eigenvalues ($\lambda_1 \ge \lambda_2$,)
computed from $-\mathsf{M}$ as defined in (\ref{equ:PG:007}),
to have $\kappa_i=\left\Vert \bm \nabla_Z \right\Vert \lambda_i$, $i=1,2$.
As the two eigenvalues of $-\mathsf{M}$ write
\begin{equation}\label{lambda}
\lambda_\pm= \frac{1}{2}\left(-\Tr(\mathsf{M}) \pm \sqrt{ (\Tr\mathsf{M})^2-4\det\mathsf{M}
}\right)
\end{equation}
 we have $\lambda_1=\lambda_+$ and $\lambda_2=\lambda_-$. Since  as well
as
\begin{equation}\label{equ:PG:800}
\kappa_1-\kappa_2={{\left\Vert \bm \nabla_Z \right\Vert(\lambda_1-\lambda_2)}}
\end{equation}
the  squared curvilinear saliency is then defined.\end{myproof}

\begin{figure}[t]
  \centering
  \setlength{\tabcolsep}{0.05cm}
  \begin{tabular}{cccc}
  {\includegraphics[width=0.22\columnwidth, height = 0.22\columnwidth]{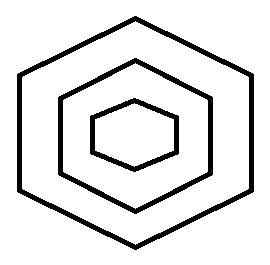}}&
  {\includegraphics[width=0.22\columnwidth, height = 0.22\columnwidth]{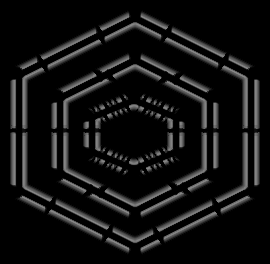}}&
  {\includegraphics[width=0.22\columnwidth, height = 0.22\columnwidth]{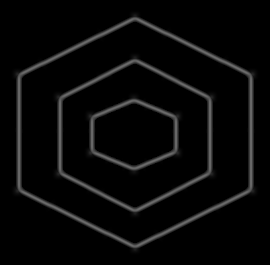}}&
  {\includegraphics[width=0.22\columnwidth, height = 0.22\columnwidth]{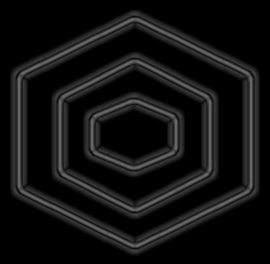}}\\
  {\includegraphics[width=0.22\columnwidth, height = 0.22\columnwidth]{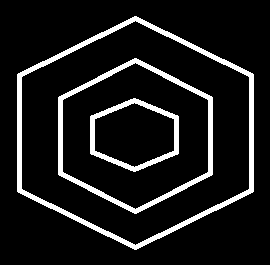}}&
  {\includegraphics[width=0.22\columnwidth, height = 0.22\columnwidth]{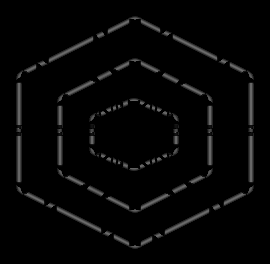}}&
  {\includegraphics[width=0.22\columnwidth, height = 0.22\columnwidth]{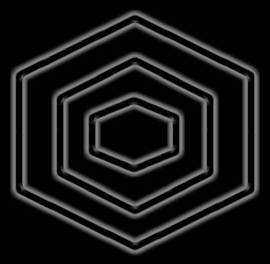}}&
  {\includegraphics[width=0.22\columnwidth, height = 0.22\columnwidth]{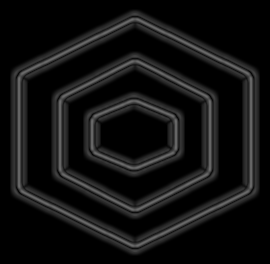}}
  \end{tabular}
  \caption{{Curvilinear saliency  of two shapes (columns 1, 5) with minimum
(2, 6), maximum (3, 7) and the difference between maximum and minimum eigenvalues
(4, 8).}}
  \label{fig:comparisonDeng}
\end{figure}

However, the rely on the highest or smallest principal
curvature alone is not adequate
for defining accurate ridges~\cite{ours}.
In Fig.~\ref{fig:comparisonDeng}, we show the different detections obtained
using the minimum or the maximum principal
curvature, as proposed by~\cite{Deng2007}.
The maximum principal
curvature provides a high response only for dark lines on a
light background, while the minimum gives the higher answers for the light
lines on a dark background.
The difference of the principal
curvatures, $\kappa_1 - \kappa_2$, improves robustness
as it responds in both settings.

\section{Image Representation}\label{sec:ima:rep}

\subsection{Proposed Curvilinear Features for Images}

Let $I(x,y)$ denote the value of the  image intensity function $I:U\subset\mathbb{R}^2\rightarrow\mathbb{R}$ at   image point $(x,y)$.
Similar to the work of~\cite{Plotz2015}, we assume the Lambertian shading model~(\ref{equ:PG:1001}). 
Let the  intensity image be treated as an intensity surface $\mathcal{I}$ defined by the vector function
\begin{equation}\label{321:EqU:002}
\mathbf{I}(x,y)=\left[ x,y,I(x,y) \right]^\top
\end{equation}
Remind that the unit normal is $\mathbf{ N}(x,y)={\mathbf{\bar N}(x,y)}/{\left\Vert\mathbf{\bar N}(x,y)\right\Vert}$ where $\mathbf{\bar N}$ is defined in~(\ref{equ:PG:300}),
and so, only depends on the depth $Z(x,y)$ and its derivatives up to order-$1$. 


We now want to detect features in the intensity surface $\mathcal{I}$ and check whether they are good candidates to be matched to curvilinear features detected 
in the depth surface  $\mathcal{D}$, w.r.t. a given camera pose. 
The key issue here is that detected features in $\mathcal{I}$ can be matched to features detected in $\mathcal{D}$ on the condition that 
both are based on measurements with the same order of derivation in  $Z(x,y)$,  in order to yield a ``compatible'' matching that {ensures  repeatablility}. 
The fact that $I$  depends on   $Z(x,y)$ and its derivatives up to order-$1$,  entails that the detection of  features in  $\mathcal{I}$ must rely
on  order-$1$ variations of the surface $\mathbf{I}(x,y)$, e.g., on its differential along some adequate direction.

Consider  a point $\mathbf{Q}=\mathbf{I}(x,y)$ on the image surface. Let   $\mathrm{d}{\mathbf{I}_\mathbf{Q}}:U\rightarrow\mathbb{R}^3$ be the differential of $\mathbf{I}$ at $\mathbf{Q}$. Given a unit direction   $\mathbf{v}=[a,b]^\top$ in the image $xy$-plane, we have $
\mathrm{d}{\mathbf{I}_\mathbf{Q}}(\mathbf{v})
=a\mathbf{I}_x+b\mathbf{I}_y=\mathtt{J}_\mathbf{I}\mathbf{v}
$
where $\mathtt{J}_\mathbf{I}$
is the Jacobian matrix of $\mathbf{I}$ and  
$\mathbf{I}_x=[1,0,I_x]^\top$ and 
$\mathbf{I}_y=[0,1,I_y]^\top$, where
\begin{equation}\label{equ:1027}
I_\star=\frac{1}{2}
\left(\Sign{\mathbf{N}\cdot\mathbf{s}}-1 \right)\left(\mathbf{N}_\star\cdot\mathbf{s}\right)
\end{equation}
$\star$ standing for either $x$ and or $y$.
It is an order-$1$ measurement of the image surface variation  at $\mathbf{Q}$ and is compatible with  our curvilinear measurements of the depth surface 
(i.e., with same order of the derivatives of $Z$).

In order to a get a scalar measurement, we define the unit vectors
$\mathbf{T}_1=\mathtt{J}_\mathbf{I}\frac{\bm\nabla_I}{\left\Vert\bm\nabla_I\right\Vert}$
and $\mathbf{T}_2$ by rotating $\mathbf{T}_1$ by $\frac{\pi}{2}$. For $\alpha=1,2$, we also define\begin{equation}\label{EQUPG:002}
\left\vert\mu_\alpha\right\vert=\left\Vert \mathrm{d}{\mathbf{I}_\mathbf{Q}}\left(\mathbf{T}_\alpha\right)
\right\Vert
\end{equation}
which is the differential of $\mathbf{I}$  along unit direction $\mathbf{T}_\alpha$ in the image plane.  
It can be easily seen that $\bm\nabla_I/{\left\Vert\bm\nabla_I\right\Vert}$ is the eigenvector of 
\begin{align}\label{eQU:PG:102}
\mathtt{J}_\mathbf{I}^\top\mathtt{J}_\mathbf{I}
=
\begin{bmatrix}
\mathbf{I}_x\cdot\mathbf{I}_x & \mathbf{I}_x\cdot\mathbf{I}_y \\
\mathbf{I}_x\cdot\mathbf{I}_y & \mathbf{I}_y\cdot\mathbf{I}_y \\
\end{bmatrix}
&=
\begin{bmatrix}
1+(I_x)^2 & I_x I_y \\
I_x I_y & 1+(I_y)^2  \\
\end{bmatrix}\notag\\
&=\mathtt{I} +\bm\nabla_I\bm\nabla_I^\top
\end{align}
associated with the largest eigenvalue $\mu_\alpha$.
It is worthy to note that the similarity between
the expression of the principal curvature computed for the depth surface, cf. (\ref{EQUPG:00}) and the formulae (\ref{EQUPG:002}).
Also, note that the matrix (\ref{eQU:PG:102}) is that of  the first fundamental
form of  $\mathcal{I}$.
Clearly, the maximum and minimum values of the quadratic form $\left\Vert\mathrm{d}{\mathbf{I}_\mathbf{Q}}(\mathbf{v})\right\Vert^2$
correspond to the two  eigenvalues of the first fundamental form matrix given in~(\ref{eQU:PG:102}).

By a similar approach to \S\ref{prop:curv:feat}, we can propose as feature the difference $\mu_1-\mu_2$
where $\mu_1\ge\mu_2$.

\begin{proposition}
Let $\mu_1,\mu_2$ be the two eigenvalues of the first fundamental form matrix
$\mathtt{J}_\mathbf{I}^\top\mathtt{J}_\mathbf{I}$ of $\mathcal{I}$,  ordered in descending
order.
Then, we have
\begin{equation}\label{equ:PG:004}
\mu_1-\mu_2=\left\Vert\bm\nabla_ I\right\Vert^2
\end{equation}
\end{proposition}
\begin{myproof}
 We can deduce the ordered eigenvalues of 
$\mathtt{I}^\mathcal{I}_\mathbf{P}
=\mathtt{J}_\mathbf{I}^\top\mathtt{J}_\mathbf{I}$
from those of   $\bm\nabla_I\bm\nabla_I^\top$, i.e., $\left\Vert\bm\nabla_
I\right\Vert^2$ and $0$, so $\mu_1=\left\Vert\bm\nabla_ I\right\Vert^2+1$
and $\mu_2=1$. Which ends the proof.
\end{myproof}

  Again, as in~\S\ref{prop:curv:feat}, we can describe the local shape of $\mathcal{I}$ at $\mathbf{Q}$
  by means of the  eccentricity of a conic, here given by
 the quadratic form $\mathbf{v}^\top\mathtt{J}_\mathbf{I}^\top\mathtt{J}_\mathbf{I}\mathbf{v}=\pm
1$. How can we interpret this conic? The first order Taylor expansion for
infinitesimal changes $(\mathrm{d}x,\mathrm{d}y)$ in the vinicity of
  $\mathbf{Q}=\mathbf{I}(x,y)$
 yields
 \begin{equation}\label{equ:003}
\mathbf{I}(x+ \mathrm{d} x, y+\mathrm{d}y) -\mathbf{I}(x,y) \approx  \mathtt{J}_\mathbf{I}
[\mathrm{d}
x,\mathrm{d} y ]^\top
\end{equation}
For any unit direction $\mathbf{v}=[a,b]^\top$
in the $xy$-plane, the quadratic form $\mathbf{v}^\top\mathtt{J}_\mathbf{I}^\top\mathtt{J}_\mathbf{I}\mathbf{v}$
returns the  linear part $g$ of growth in arc length
from $\mathbf{I}(x,y)$ to $\mathbf{I}(x+a,y+b)$. Therefore, we have
\begin{equation}\label{eQu:PG:002}
g^2=\left\Vert\mathrm{d}{\mathbf{I}_\mathbf{Q}}((\mathrm{d}x,\mathrm{d}y)\right\Vert^2
=\mathbf{v}^\top\mathtt{J}_\mathbf{I}^\top\mathtt{J}_\mathbf{I}\mathbf{v}
\end{equation}

An important remark that we highlight here and not mentioned in~\cite{Plotz2015} is the following one. 
The AVG feature defined in~(\ref{equ:PG:103}) is actually the Frobenius norm of the Jacobian matrix $\mathtt{J}_\mathbf{N}$ of the map $\mathbf{N}(x,y)$, see~(\ref{EQU:432:001}),
up to  constant $\gamma$.  
Clearly, this describes the second-order behaviour of the surface  $\mathcal{M}$ relative to the  normal at one of its points in the immediate vicinity of this point. 
Using the result in~(\ref{equ:PG:300}), (\ref{EQU:432:001}) and (\ref{EQU:432:002}), we can claim that the extracted feature in the virtual image only depends on $X, Y,Z$ 
and their derivatives up to order-$2$. 
It is consistent (regarding the considered orders of the derivatives of $X,Y,Z$) with the feature $\left\Vert\bm\nabla_ I\right\Vert = \sqrt{{I_x^2+I_y^2}}$ 
detected in the intensity image, where $I_\star$, with $\star\in\{1,2\}$ is given in~(\ref{equ:1027}).

\subsection{Multi-Curvilinear Saliency (MCS)}\label{MCS2}

Multi-scale helps to detect important structures as well as small details.
In consequence, in this paper, we compute the curvilinear saliency images in a multi-scale space.
To build the scale pyramid, an edge-preserving smoothing approach, named anisotropic diffusion filter~\cite{Paris2009}, is used in order to avoid oversmoothing.
In fact, this filter tries to separate the low frequency components (i.e, sharp edges) from the high frequency components (i.e., textures) 
by preserving the largest edges in an image.

Contrary to depth images which represent textureless 3D shapes, intensity images are composed of shape and texture components.
Consequently, the curvilinear saliency (CS) estimated from intensity images is affected by the textured regions.
Our idea is to put forward the assumption that multi-scale analysis can discriminate between keypoints 
(those with high CS value in the image) due to shape and keypoints due to texture.
At a coarse level, edges detected are reliable but with a poor localization and they miss small details.
At a fine level, details are preserved, but detection suffers greatly from clutters in textured regions.
In addition, the CS 
values of small details and textures are high in the coarse level,
whereas these values become lower in the finest levels.
To combine the strengths of each scale, the CS 
value of each pixel over $n$ scales is analyzed.
If this value in all scales is higher than a threshold $T$, the maximum curvilinear saliency ($\operatorname*{MCS}$) value of this pixel over all scales is then kept.
This threshold is a function of the number of the smoothed images, $n$, (i.e., $T = e^{-n}$: when $n$ is small, then $T$ is a big value and vice versa).
However, if the CS 
value is lower than $T$ in one level, it is considered as a point that belongs to a texture (or a small detail) point,
thus it is removed from the final multi-scale curvilinear saliency, $\operatorname*{MCS}$, image.
Adding this multiscale step should help to reduce the impact of the texture on the point of interest detection. 
However, in the next section, we propose to introduce the principle used for estimating focus map 
in order to increase the robustness to the background and to the presence of the texture. 

\section{Robustness to background and to texture}\label{sec:robustness}

Before introducing the proposed improvement, we briefly present existing works about texture detection and, in particular, about focus curve estimation. 
\subsection{Extraction of texture: state of the art}

Various methods, such as~\cite{Aujol2006,Zhang2014,Karacan2013,Jeon2014} have been proposed for extracting the texture from a natural image.
In these approaches, a given image is separated into two components while  preserving edges. In~\cite{Aujol2006}, Aujol et al. proposed a variational model based on total-variation (TV) energy for extracting the structural part. 
In~\cite{Zhang2014}, the authors proposed an algorithm in the field of scale space theory. 
This technique is a rolling guidance method based on an associated depth image to automatically refine the scale of the filtering in order to preserve edges. 
A structure-preserving image smoothing approach is introduced in~\cite{Karacan2013}.
This method locally analyzes the second order feature statistics of a patch around a pixel.
The algorithm used a 7-dimensional feature vector that contains intensity, orientation and pixel coordinates.
Finally, under a condition that the images contain smooth illumination changes and Lambertian surfaces,
\cite{Jeon2014} proposed an intrinsic image decomposition model which explicitly determines a separate texture layer, as well as the shading layer and the reflectance layer.
The method is based on surface normal vectors generated from an RGB-D image.
All these works first smooth the intensity image as a pre-processing stage and then extracting the shape from that image relying on  prior knowledge.
And, to sum up, most of these methods for structure-texture decomposition are analogous to the classical signal processing low pass-high pass filter decomposition.
However, even if it is correct to consider that the structure part of an image contains strong edges, the texture can also contain medium and high frequencies and the texture can only be partially removed.
Another possibility is to consider focusness, related to the degree of focus.

Usually focusness is defined as inversely proportional to the degree of blur (blurriness)~\cite{Jiang2013}.
It is a very valuable tool for depth recovery~\cite{Zhuo2011} but also for blur magnification, or for image quality assessment.
Blur is usually measured in regions containing edges, since edges would appear in images as blurred luminance transitions of unknown blur scale~\cite{Elder1998}. Then the estimation of the blur can be propagated to the rest of image.
Since blur occurs for many different causes, this task is challenging and, in the literature, many methods on focus map estimation have been proposed.
In~\cite{Elder1998}, the authors identify blur as focal blur, induced by  finite depth of field, as penumbra blur or shading blur and then estimate the blur scale. 
In the context of matting method~\cite{Zhuo2011}, the blur ratio for every pixel corresponds to the ratio between the gradients of the input image and the re-blurred images.
In~\cite{Wang2016}, the authors use the K nearest neighbors (KNN) matting interpolation under the assumption that depth is locally uniform.
However, blurring can also appear with edges caused by shadows and glossy highlights can also produce error in focus estimation.
To remove errors induced by these other sources of blur, \cite{Zhuo2011}  used a cross bilateral filtering and estimation of sharpness bias and~\cite{Jiang2013} used multi-scale.
In the field of saliency detection, another way to estimate the amount of blur consists in computing the defocus blur between two Difference of Gaussian (DoG) images 
in multi-scale levels~\cite{Jiang2013}.
An optimization algorithm is used by minimizing the difference between the blurriness of a pixel and the weighted average of blurriness of neighboring pixels.
In a Markov Random Field (MRF) formulation, a local contrast prior based on comparing local contrast and local gradient is also introduced in~\cite{Tai2009}.
In~\cite{Kumar2015}, the authors propose to use the ratio between principle components, that is a weighted mixture of the spectral components.
Moreover, the weights are proportional to the energy in the spectral component.
Some algorithms also use the analysis of localized Fourier (Gabor filtering) spectrum~\cite{Zhu2013}.
In addition, smoothness constraints and image color edge information are taken into account to generate a defocus map for trying to preserve discontinuities 
on the transitions between objects.

Following all these aforementioned approaches, we can find that most of the existing algorithms~\cite{Zhuo2011,Jiang2013,Kumar2015}
depend on measuring the blur amount using the ratio between the edges in two different scale levels (i.e., the original image and the re-blurred image).
In consequence, we propose to use the ratio between the two curvilinear saliency images that contain robust edges in different scales to determine the blur amount based on the methods developed in~\cite{Zhuo2011}. 
For the multi-scale aspect, our approach is inspired by the principles explained in~\cite{Jiang2013}.

\subsection{Removing background with focus curves: state of the art}

Based on the mapping between the depth of a point light source and the focus level of its image,
Shape From Defocus (SFD) approaches recover the 3D shape of a scene from focused images that represent the focus level of each point in the scene~\cite{Pentland1987}.
We can also notice that the focus (defocus) maps can be also used as an alternative for depth map, like in existing Adobe tools~\cite{Zhuo2011,Jiang2013,Zhu2013}.
Consequently, it seems interesting to introduce what we call the detection of ``focus curves'' that capture blurriness in images.
More precisely, focus curves mean that we estimate the scale of blur at the curvilinear saliency feature of the original image and we suppose that these features should be only related to discontinuities.

\begin{figure}[h!]
\centering
\includegraphics[width=0.9\columnwidth]{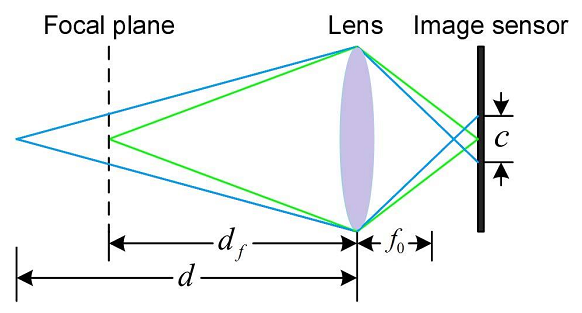}
\caption{A thin lens model for image blur as proposed in~\cite{Zhuo2011}. }
\label{BlurModel}
\end{figure}

Focal blur occurs when a point is out of focus, as illustrated in Fig~\ref{BlurModel}.
When the point is at the focus distance, $d_f$, from the lens, all the rays from it converge to a sharp single sensor point.
Otherwise, when $d \neq d_f$, these rays generate a blurred region in the sensor area.
The blur pattern generated by this way is called the circle of confusion (CoC) whose diameter is denoted $c$.

\begin{figure*}
\centering
\includegraphics[width=2\columnwidth]{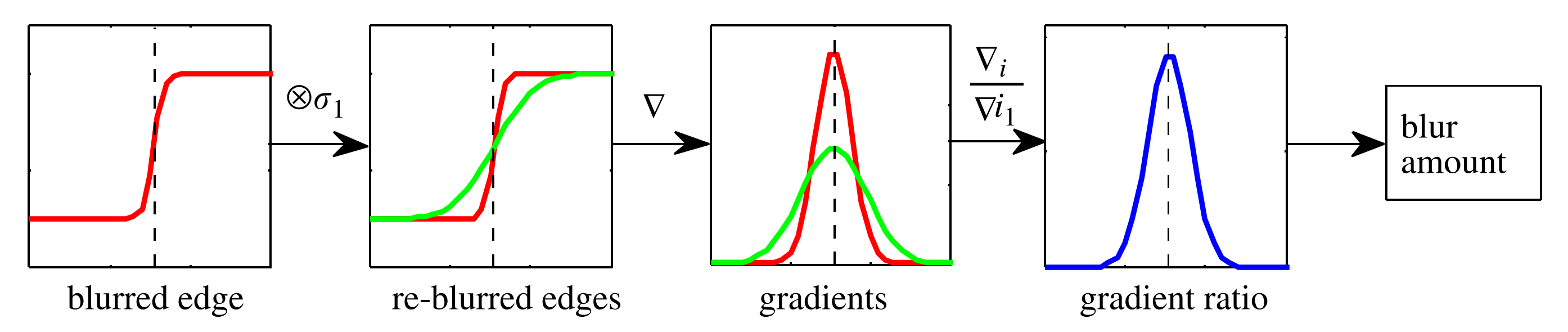}
\caption{Overview of our blur estimation approach: here, $\bigotimes$ and $\nabla$ are respectively the convolution and the gradient operators, $\sigma_1$ is the standard deviation of a re-blurring Gaussian function.
The black dash line denotes the edge location as proposed in~\cite{Zhuo2011}.}
\label{Blurratio}
\end{figure*}

In~\cite{Soonmin2007,Zhuo2011,Jiang2013}, the defocus blur can be modeled as a convolution of a sharp image with the point spread function (PSF) as shown in Fig.~\ref{Blurratio}.
The PSF is usually approximated by a Gaussian function $g(x, \sigma)$, where the standard deviation $\sigma \propto c$ measures the blur amount and is proportional to the diameter of the CoC:
\[c = \frac{|d-d_f|}{d}\frac{f}{d-f},\]
where $d, d_f, f$ are focus distance, defocus distance and focal length respectively as shown in Fig.~\ref{BlurModel}.
A blurred edge $i(x)$ is then given by
\begin{align}
\label{edgeblur}
i(x)=f(x)\otimes g(x,\sigma)
\end{align}
where $f(x)=Au(x)+B$ is an ideal edge where $u(x)$ is the step function. 
The terms $A$ and $B$ corresponds to the amplitude and the offset of the edge, respectively. Note that the edge is located at $x=0$.

In~\cite{Zhuo2011}, the blur estimation method was described for 1D case. The gradient of the re-blurred edge is:
\begin{align}
\label{edgeblur2}
\nabla i_{1}(x) =& \nabla (i(x)\otimes g(x,\sigma_0))\nonumber\\
= & \nabla ((Au(x)+B) \otimes g(x,\sigma) \otimes g(x,\sigma_0))\nonumber\\
= & \frac{A}{\sqrt{2\pi (\sigma + \sigma_0)}} \exp \left(-\frac{x^2}{2(\sigma^2 + \sigma_0^2)}\right)
\end{align}
where $\sigma_0$ is the standard deviation of the re-blur Gaussian kernel. 
Thus, the gradient magnitude ratio between the original and the re-blurred edges is:
\begin{equation}
\begin{split}
\label{ratiogradient}
R &= \frac{\mid \nabla i(x)\mid}{\mid \nabla i_{1}(x)\mid} \\
&= \sqrt{\frac{\sigma^2 + \sigma_0^2}{\sigma^2}}\exp -\left(\frac{x^2}{2(\sigma^2)}-\frac{x^2}{2(\sigma^2 + \sigma_0^2)}\right).
\end{split}
\end{equation}

It can be proved that the ratio is maximum at the edge location ($x=0$) and the maximum value is given by:

\begin{equation}
\label{ratiogradient2}
R = \sqrt{\frac{\sigma^2 + \sigma_0^2}{\sigma^2}}
\end{equation}

Finally, given the maximum value $R$ at the edge locations, the unknown blur amount $s$ can be calculated using:

\begin{equation}
\label{ratiogradient3}
s = \frac{\sigma_0}{\sqrt{R^2-1}}
\end{equation}

\subsection{Focus curves based on Curvilinear Saliency -- Multi Focus Curves (MFC)}

We suppose that using focus can help to remove the background and using multiscale can help to reduce the influence of the texture in the same way as in section~\ref{MCS2}.
So, we propose to use the curvilinear saliency computation instead of the edge response to estimate the focus curves of an input image.
In addition, we propose to estimate focus curves in multi-scales rather than in one scale as proposed in~\cite{Zhuo2011} to have scale invariant focus salient curves.
In addition, we combine all information gotten from different blurring scales.

Assume the original pixel in an image is blurred as $F(x,y)= {I}(x,y)\otimes g(x,y,\sigma)$. Thus to get  the curvilinear saliency, the structure tensor is calculated as:
\begin{align}
\label{Hessian}
\mathsf{S_T} =& f\left(\nabla \left(\left({I}(x,y) \otimes g(x,y,\sigma)\right)\right)\right)\nonumber\\
    =&\left( \begin{array}{cc}
     {I}_{x}^2  &{I}_{x}I_y\\
     {I}_{x}I_y  &{I}_{y}^2
 \end{array} \right)\otimes g(x,y,\sigma)
\end{align}

If the Hessian matrix is expressed with eigenvectors and eigenvalues, we obtained:
\begin{align}
\label{HessianCompse}
\mathsf{H} =\left(\left[ \begin{array}{cc}
     {e_1}  &{e_2}
 \end{array} \right] \left[ \begin{array}{cc}
     {\lambda}_{1}  &{0}\\
     {0}  &{\lambda}_{2}
 \end{array} \right] \left[ \begin{array}{c}
     {e_1^T} \\
     {e_2^T}
 \end{array} \right] \right) \otimes g(x,y,\sigma)
 \end{align}

 \begin{align}
 \mathsf{H} = \left( (\lambda_1-\lambda_2) e_1e_1^T + \lambda_2(e_1e_1^T+e_2e_2^T)\right)\otimes g(x,y,\sigma)
\end{align}
The curvilinear saliency can be described as:
\begin{align}
\label{CSl}
CS = (\lambda_1 -\lambda_2) \otimes g(x,y,\sigma)
\end{align}
In particulary, the curvilinear saliency can be directly computed as:
\begin{align}
\label{CSgraident}
CS = \alpha ((I_{x}^2+I_{y}^2)) \otimes g(x,y,\sigma)
\end{align}
The re-blurred curvilinear saliency image, named $CS_i$, in multi-scale can then be defined as:
\begin{align}
\label{CSmultiscale}
CS_i = \alpha ((I_{x}^2+I_{y}^2)) \otimes g(x,y,\sigma) \otimes g(x,y,\sigma_i),\enspace i=1,2,...,n
\end{align}
where $n$ is the number of scales.

Consequently, the ratio between the original and re-blurred curvilinear saliency is:
\begin{align}
\label{CSmultiscale2}
R_i = \frac{CS_i}{CS} = \frac{\sigma^2 + \sigma_{i}^2}{\sigma^2}\exp -\left(\frac{x^2+y^2}{2(\sigma^2)}-\frac{x^2+y^2}{2(\sigma^2 + \sigma_{i}^2)}\right)
\end{align}
Within the neighborhood of an pixel, the response reaches its maximum when $x = 0$ and $y =0$, thus:
 \begin{align}
\label{CSmultiscalezero}
R_i\mid_{0} = \frac{CS_i}{CS} = \frac{\sigma^2 + \sigma_{i}^2}{\sigma^2} = 1+\frac{\sigma_{i}^2}{\sigma^2}
\end{align}
Finally, given the maximum value $R_i$ in each scale level, the unknown blur amount $s_i$ can be calculated using
\begin{equation}
\label{ratiogradient4}
s_i = \frac{\sigma_i}{\sqrt{R_i\mid_{0}-1}},
\end{equation}

For $n$ scales, we compute $n-1$ focus curve scales by using the ratio between curvilinear saliency of the coarse level (i.e., the original image) and the next scale levels.
By following the same remarks as in section~\ref{MCS2}, we define 
Multi Focus Curves (MFC) that correspond to the fusion of all the focus curves into one map by keeping only the pixels that have
a high focus value in all the $n-1$ scales (i.e., a high value means a value bigger than $T=e^{-n}$, chosen in the same way as section~\ref{MCS2}).
If the pixel has a high value at all scales, the maximum value of the scale of blur is taken into account to build the final multi-scale curve map :
\begin{align}
\label{ratiogradientMerge}
MFC = \frac{1}{\arg \max_{i}\;(s_i)}.
\end{align}

In conclusion, the highest values of the estimated $MFC$ indicate edges that have low blurring (i.e., sharp edges).
On the contrary, low values indicate ones that have high blurring.
Consequently, we expect that focus curves highlight salient curvilinear saliency in images that are approximately similar to the detected curvilinear saliency features in depth images.

\section{Experiments for feature detection}\label{exp}
\subsection{Comparison with existing methods}

One of our most important objectives in this work was to introduce a detector that is more repeatable between 2D images and 3D models than classical detectors in the litterature. 
In consequence, we compare the features detected on 3D models with the proposed curvilinear saliency detector with features detected on real images with these three 2D detectors:
Image Gradient (IG), Multi-scale Curvilinear Saliency (MCS) and  Multi-scale Focus Curves (MFC). 
In addition, we measure the repeatability between the two others 3D model detectors, i.e. Average Shading Gradient (ASG)~\cite{Plotz2015} and Hessian Frobenius Norm (HFN), 
and the same three 2D detectors. And then, we compared MFC and MCS with nine classical 2D detectors:
\begin{enumerate}
 \item \textbf{Edge detectors}:  (\textit{i}) Sobel,
  (\textit{ii}) Laplacian of Gaussian (Log),
  (\textit{iii}) Canny~\cite{canny1986} and
  (\textit{iv}) Fuzzy logic technique~\cite{Kiranpreet2010};
  \item \textbf{Corner detectors}:  (\textit{v}) Harris detector based on auto-correlation  analysis and
 (\textit{vi}) Minimum Eigenvalues detector based on analysis of the Hessian matrix~\cite{Tomasi1994};
 \item \textbf{Multi-scale detectors}: (\textit{vii}) SIFT, Scale Invariant Feature Transform~\cite{Lowe2004}, that uses the analysis of difference of Gaussian,
 (\textit{viii}) SURF, Speeded Up Robust Features, a multi-scale technique based on the Hessian matrix~\cite{Bay2008} and
  (\textit{ix}) a multi-scale Principal Curvature Image (PCI) detector~\cite{Deng2007}.
\end{enumerate}

\subsection{Evaluation criteria}
The eleven 2D detectors are evaluated with two \noindent\textbf{measures}:
\begin{enumerate}
 \item Intersection percentage (IP): the probability that a 2D intensity-based key feature can be found close to those extracted in a depth image~\cite{ours}.
 \item Hausdorff Distance (HD): the classical measurement is defined for two point sets $A$ and $B$ by: 
 \[HD(A,B)=\max\left(h(A,B), h(B,A)\right), \]
 where $h(A,B) = \max\limits_{a\in A} \min\limits_{b\in B} \parallel a-b \parallel$. The lowest the distance, the most similar the two sets.
\end{enumerate}

 \subsection{Datasets}

Two datasets are evaluated:
\begin{enumerate}
 \item \textit{Web collection}: we have collected $10$ objects and $15$ real images of each object on the web by choosing views as close as possible to the views used for the generation of the depth images.
Moreover, to highlight the robustness of the approach to different acquisition conditions, many real images of a similar model are taken.
\item \textit{PASCAL3D+} dataset~\cite{PASCAL3D}: it is used in order to assess scalability. it contains real images corresponding to $12$ rigid objects categories.
We have computed average results for all non occluded objects in each category, i.e. around $1000$ real images and $3$ or more reference models per category.
The real images are acquired under different acquisition conditions (e.g., lighting, complex background, low contrast).
We have rendered the depth images of the corresponding 3D CAD model using the viewpoint information from the dataset.
Only non-occluded and non-truncated objects in the real images were used. Furthermore, we choose 3D textureless objects (available online \footnote{\tiny{\url{http://tf3dm.com/}}}),
\end{enumerate}
For all the tested 3D models, we have rendered depth images using \emph{MATLAB 3D Model Renderer} \footnote{\tiny{\url{http://www.openu.ac.il/home/hassner/projects/poses/}}}.

\subsection{Analysis of the results}

As shown in tables~\ref{DTLSampleIP} and~\ref{IPPASCAL3D}, and as expected, the proposed approach using focus curves based on curvilinear saliency, named MFC, 
is able to find the highest number of features in the intersection with the features detected on real images captured under different textures and lighting conditions.
More precisely, MFC obtains an average mean intersection percentage greater than $56\%$ whereas for MCS and PCI, it is respectively greater than $50\%$ and $44\%$, 
for the web collection dataset.
With the PASCAL+3D dataset, MFC also yields the highest mean average IP among all the tested detectors that is $46\%$.

\begin{table}[htb]
\begin{center}
\setlength{\tabcolsep}{0.01cm}
\begin{tabular}{|>{\footnotesize}c|>{\footnotesize}c|>{\footnotesize}c|>{\footnotesize}c||>{\footnotesize}c|>{\footnotesize}c|>{\footnotesize}c|>{\footnotesize}c||>{\footnotesize}c|>{\footnotesize}c|>{\footnotesize}c|>{\footnotesize}c|}
\hline
\cellcolor{red!25}Methods   &{\cellcolor{red!25}MFC} &{\cellcolor{red!25}MCS} &{\cellcolor{red!25}PCI} &{\cellcolor{red!25}MinEig} &{\cellcolor{red!25}Harris} &{\cellcolor{red!25}SIFT} &{\cellcolor{red!25}SURF} &{\cellcolor{red!25}Sobel} &{\cellcolor{red!25}Canny} &{\cellcolor{red!25}LOG}    &{\cellcolor{red!25}Fuzzy} \\ \hline
 \cellcolor{green!25}Car      &\cellcolor{blue!25}{59}   &50  &46  &08  &04  &03  &03  &10  &18  &11  &05\\ \hline
 \cellcolor{green!25}Shoe     &\cellcolor{blue!25}{38}   &31  &31  &02  &03  &10  &01  &04  &04  &05  &02\\ \hline
 \cellcolor{green!25}Plane    &\cellcolor{blue!25}{58}   &55  &38  &06  &04  &10  &03  &18  &21  &21  &14 \\ \hline
 \cellcolor{green!25}T-Rex    &\cellcolor{blue!25}{66}   &64  &59  &09  &06  &02  &05  &16  &18  &20  &12 \\ \hline
 \cellcolor{green!25}Elephant &\cellcolor{blue!25}{37}   &32  &32  &03  &03  &05  &03  &06  &08  &06  &04 \\ \hline
 \cellcolor{green!25}Fhydrant &\cellcolor{blue!25}{56}   &51  &42  &06  &04  &02  &09  &09  &14  &13  &06 \\ \hline
 \cellcolor{green!25}Jeep     &\cellcolor{blue!25}{69}   &62  &58  &05  &05  &05  &06  &09  &15  &11  &06 \\ \hline
 \cellcolor{green!25}Mug      &\cellcolor{blue!25}{57}   &54  &50  &02  &03  &04  &03  &08  &12  &07  &08 \\ \hline
 \cellcolor{green!25}Teddy    &\cellcolor{blue!25}{44}   &39  &32  &04  &05  &09  &04  &07  &14  &08  &07 \\ \hline
 \cellcolor{green!25}Pistol   &\cellcolor{blue!25}{69}   &67  &61  &09  &09  &09  &04  &13  &23  &14  &07 \\ \hline

 \end{tabular}
\end{center}
\caption{Mean Intersection Percentage (IP) (\emph{higher is better}) of all depth images rendered from different viewpoints and 
all real images captured under different textures and lighting for the web collection with the proposed method (MFC), the method (MCS) and $9$ tested detectors. }
\label{DTLSampleIP}
\end{table}

\begin{table}[htb]
\begin{center}
\setlength{\tabcolsep}{0.02cm}
\begin{tabular}{|>{\footnotesize}c|>{\footnotesize}c|>{\footnotesize}c|>{\footnotesize}c||>{\footnotesize}c|>{\footnotesize}c|>{\footnotesize}c|>{\footnotesize}c||>{\footnotesize}c|>{\footnotesize}c|>{\footnotesize}c|>{\footnotesize}c|}
\hline
\cellcolor{red!25}Methods   &{\cellcolor{red!25}MFC} &{\cellcolor{red!25}MCS} &{\cellcolor{red!25}PCI} &{\cellcolor{red!25}MinEig} &{\cellcolor{red!25}Harris} &{\cellcolor{red!25}SIFT} &{\cellcolor{red!25}SURF} &{\cellcolor{red!25}Sobel} &{\cellcolor{red!25}Canny} &{\cellcolor{red!25}LOG}    &{\cellcolor{red!25}Fuzzy} \\ \hline
  \cellcolor{green!25}plane       &\cellcolor{blue!25}{55} &50 &37 &15 &09 &08  &13  &10 &13 &11 &10\\ \hline
  \cellcolor{green!25}bicycle     &\cellcolor{blue!25}{69} &61 &57 &25 &08 &16 &24  &13 &15 &18 &14\\ \hline
  \cellcolor{green!25}boat        &\cellcolor{blue!25}{42} &36 &28 &09  &10  &06  &10  &09 &14 &11 &09\\ \hline
  \cellcolor{green!25}bus         &\cellcolor{blue!25}{31} &24  &17 &05  &06 &02 &04  &04 &06 &04 &04\\ \hline
  \cellcolor{green!25}car         &\cellcolor{blue!25}{44} &41  &24 &08  &08  &03 &06  &16 &18 &14 &13\\ \hline
  \cellcolor{green!25}chair       &\cellcolor{blue!25}{56} &52  &43 &16  &08  &09  &16  &24 &20 &22 &19\\ \hline
  \cellcolor{green!25}table       &\cellcolor{blue!25}{40} &38 &19 &06  &05 &04  &08  &11 &12 &11 &07\\ \hline
  \cellcolor{green!25}train       &\cellcolor{blue!25}{31} &28 &14  &06  &07  &03  &05  &08 &07 &04 &06\\ \hline
\end{tabular}
\end{center}
\caption{Mean Intersection Percentage (IP) (\emph{higher is better}) of all depth images rendered from different viewpoints and 
all real images captured under different textures and lighting for the PASCAL3D+ with the proposed method (MFC), the method (MCS) and $9$ tested detectors. }
\label{IPPASCAL3D}
\end{table}

In addition, as shown in~\ref{DTLSampleHD} and~\ref{HDPASCAL3D}, the average Hausdorff Distance (HD) with MFC is less than $24$ and with MCS, less than $32$.
On the contrary, the other detectors do not reach high repeatability scores. 

All these quantitative results support that MFC is able to detect curvilinear saliency features that are more repeatable 
between an intensity image and its corresponding depth image than the state of the art.

\begin{table}[htb]
\begin{center}
\setlength{\tabcolsep}{0.02cm}
\begin{tabular}{|>{\footnotesize}c|>{\footnotesize}c|>{\footnotesize}c|>{\footnotesize}c||>{\footnotesize}c|>{\footnotesize}c|>{\footnotesize}c|>{\footnotesize}c||>{\footnotesize}c|>{\footnotesize}c|>{\footnotesize}c|>{\footnotesize}c|}
\hline
\cellcolor{red!25}Methods  &{\cellcolor{red!25}MFC} &{\cellcolor{red!25}MCS} &{\cellcolor{red!25}PCI} &{\cellcolor{red!25}MinEig} &{\cellcolor{red!25}Harris} &{\cellcolor{red!25}SIFT} &{\cellcolor{red!25}SURF} &{\cellcolor{red!25}Sobel} &{\cellcolor{red!25}Canny} &{\cellcolor{red!25}LOG}   &{\cellcolor{red!25}Fuzzy} \\ \hline
\cellcolor{green!25}Sequences&\cellcolor{blue!50} {HD} &\cellcolor{blue!50} {HD}  &\cellcolor{blue!50} {HD}  &\cellcolor{blue!50} {HD}  &\cellcolor{blue!50} {HD}  &\cellcolor{blue!50} {HD}  &\cellcolor{blue!50} {HD} &\cellcolor{blue!50} {HD}  &\cellcolor{blue!50} {HD}  &\cellcolor{blue!50} {HD}  &\cellcolor{blue!50} {HD}\\ \hline
\cellcolor{green!25}Car      &\cellcolor{yellow!25}{21} &29  &40  &57   &77  &85  &71  &48 &46 &47 &49\\ \hline
\cellcolor{green!25}Shoe     &\cellcolor{yellow!25}{34} &52  &67  &102  &106 &111 &108 &71 &71 &71 &71\\ \hline
\cellcolor{green!25}Plane    &26 &\cellcolor{yellow!25}{23}  &19  &37   &43  &46  &47  &26 &26 &24 &24\\ \hline
\cellcolor{green!25}T-Rex    &20 &\cellcolor{yellow!25}{17}  &25  &41   &100 &143 &46  &28 &28 &32 &22\\ \hline
\cellcolor{green!25}Elephant &\cellcolor{yellow!25}{21} &41  &55  &80   &91  &114 &74  &57 &58 &57 &57\\ \hline
\cellcolor{green!25}Fhydrant &\cellcolor{yellow!25}{15} &23  &35  &62   &86  &74  &67  &38 &37 &36 &42\\ \hline
\cellcolor{green!25}Jeep     &\cellcolor{yellow!25}{29} &31  &42  &70   &67  &74  &89  &47 &47 &46 &47\\ \hline
\cellcolor{green!25}Mug      &\cellcolor{yellow!25}{35} &56  &65  &129  &133 &134 &145 &72 &76 &75 &75\\ \hline
\cellcolor{green!25}Teddy    &\cellcolor{yellow!25}{19} &24  &31  &72   &69  &77  &101 &47 &44 &47 &47\\ \hline
\cellcolor{green!25}Pistol   &18 &\cellcolor{yellow!25}{16}  &26  &34   &96  &44  &73  &30 &65 &29 &26\\ \hline
 \end{tabular}
\end{center}
\caption{Mean Hausdorff Distance (HD) (\emph{lower is better})of all depth images rendered from different viewpoints and 
all real images captured under different textures and lighting for the web collection with the proposed method (MFC), the method (MCS) and $9$ tested detectors. }
\label{DTLSampleHD}
\end{table}

\begin{table}[htb]
\begin{center}
\setlength{\tabcolsep}{0.02cm}
\begin{tabular}{|>{\footnotesize}c|>{\footnotesize}c|>{\footnotesize}c|>{\footnotesize}c||>{\footnotesize}c|>{\footnotesize}c|>{\footnotesize}c|>{\footnotesize}c||>{\footnotesize}c|>{\footnotesize}c|>{\footnotesize}c|>{\footnotesize}c|}
\hline
\cellcolor{red!25}Method&{\cellcolor{red!25}MFC} &{\cellcolor{red!25}MCS} &{\cellcolor{red!25}PCI} &{\cellcolor{red!25}MinEig} &{\cellcolor{red!25}Harris} &{\cellcolor{red!25}SIFT} &{\cellcolor{red!25}SURF} &{\cellcolor{red!25}Sobel} &{\cellcolor{red!25}Canny} &{\cellcolor{red!25}LOG}   &{\cellcolor{red!25}Fuzzy} \\ \hline
  \cellcolor{green!25}plane    &\cellcolor{yellow!25}{47}  &48  &59  &61  &63  &68  &73  &68   &65  &69  &71\\ \hline
  \cellcolor{green!25}bicycle  &\cellcolor{yellow!25}{71}  &75  &79  &90  &101 &93  &100 &83   &84  &82  &87\\ \hline
  \cellcolor{green!25}boat     &\cellcolor{yellow!25}{62}  &68  &75  &79  &77  &87  &76  &75   &71  &78  &76\\ \hline
  \cellcolor{green!25}bus      &\cellcolor{yellow!25}{106} &110 &117 &128 &123 &131 &127 &121  &118 &122 &123\\ \hline
  \cellcolor{green!25}car      &\cellcolor{yellow!25}{80}  &85  &98  &102 &100 &113 &108 &89   &88  &94  &97\\ \hline
  \cellcolor{green!25}chair    &\cellcolor{yellow!25}{62}  &64  &78  &84  &96  &94  &86  &88   &91  &86  &92\\ \hline
  \cellcolor{green!25}table    &\cellcolor{yellow!25}{84}  &85  &96  &117 &118 &118 &111 &117  &114 &116 &120\\ \hline
  \cellcolor{green!25}train    &\cellcolor{yellow!25}{101} &108 &121 &126 &123 &133 &127 &125  &129 &129 &122\\ \hline
\end{tabular}
\end{center}
\caption{Mean Hausdorff Distance (HD) (\emph{lower is better}) of all depth images rendered from different viewpoints and 
all real images captured under different textures and lighting for the PASCAL3D+ with the proposed method (MFC), the method (MCS) and $9$ tested detectors. }\label{HDPASCAL3D}
\end{table}

\begin{figure}[htb]
\centering
\begin{tabular}{c}
  \includegraphics[width=0.9\columnwidth, height = 0.6\columnwidth]{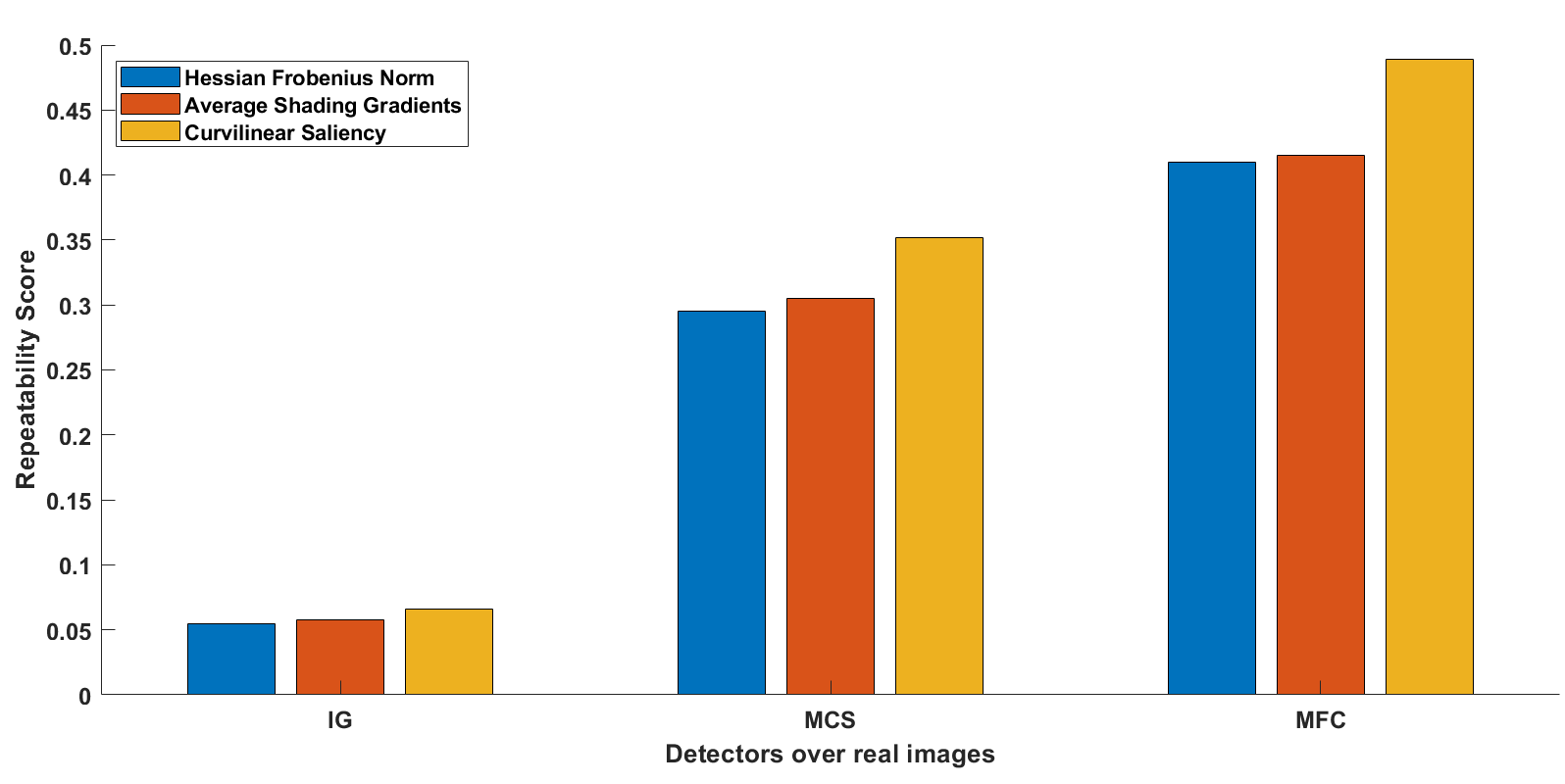}\\
  (a)\\
  \includegraphics[width=0.9\columnwidth, height = 0.6\columnwidth]{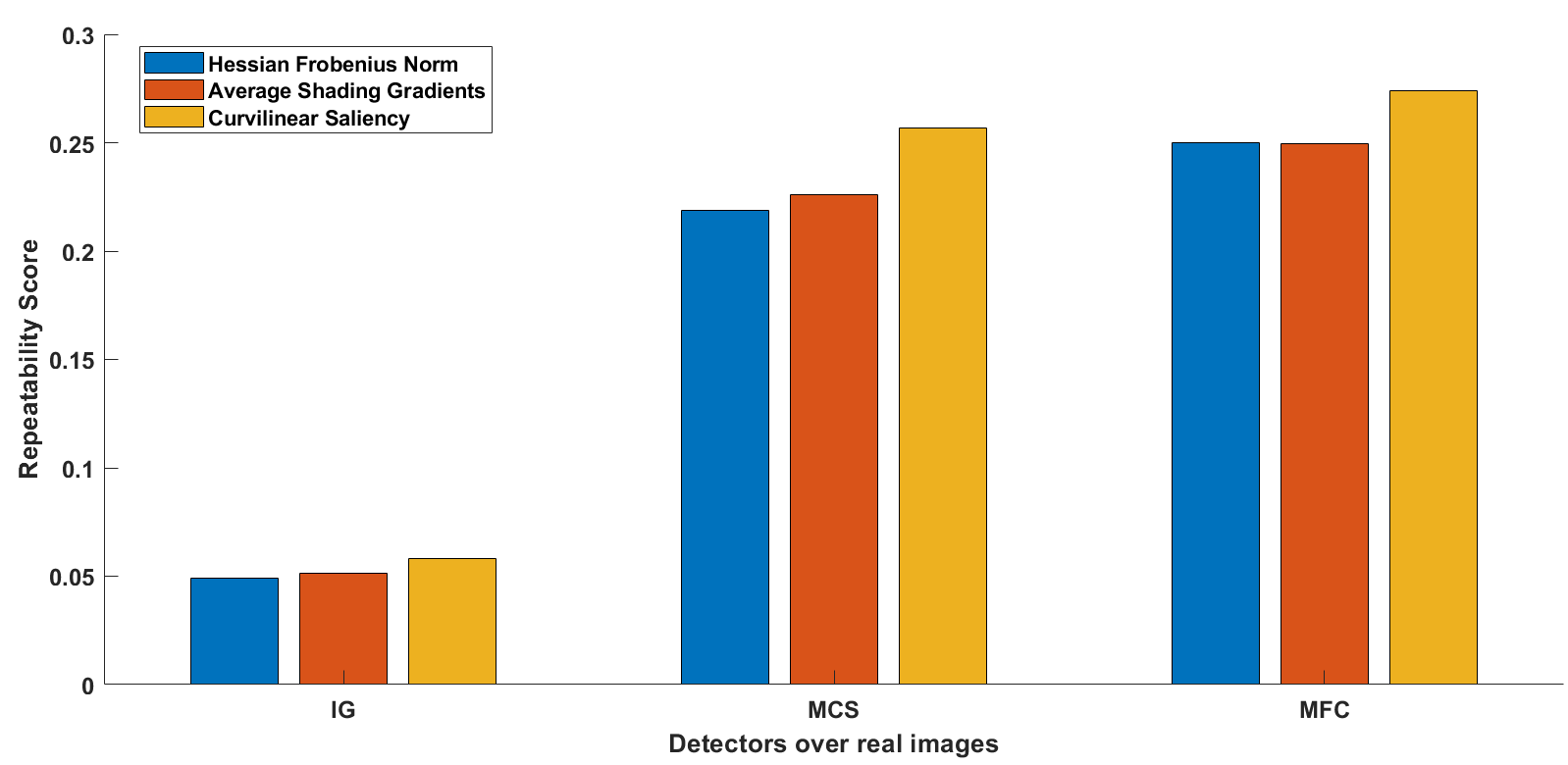}\\
  (b)\\
  \end{tabular}
  \caption{Average repeatability percentages for two examples of 3D models of the PASCAL3D+ dataset~\cite{PASCAL3D}: car (a) and sofa (b) models.}
  \label{Rep_res_Pascal}
\end{figure}

In the rest of this section, we illustrate the results for the most significant dataset, the PASCAL3D+, in order to avoid redundancy in the explanations. 
In Fig.~\ref{Rep_res_Pascal}, two examples of results obtained for the PASCAL3D+ dataset~\cite{PASCAL3D} are given.
More precisely, the repeatability percentage between the three comparable 3D detectors, i.e. MFC, MSC and Image Gradient (IG), and 
the three comparable 2D detectors, Hessian Frobenius Norm, Average Shading Gradient and 
CS, is presented. 
These results highlight that image gradients are effected by texture.
Moreover, MCS improves the repeatability between depth images and real images, compared to IG.
And, as expected, MFC still yields the best repeatability scores. 
Among the detectors used for depth images, the Curvilinear Saliency detector yields the best repeatability scores between the three intensity-based 2D detectors. 
In conclusion, using CS with MFC gives the best repeatability among all the other possible combinations. 

\begin{figure}[htb]
  \centering
  \begin{tabular}{ccc}
  \includegraphics[width=0.29\columnwidth, height = 0.22\columnwidth]{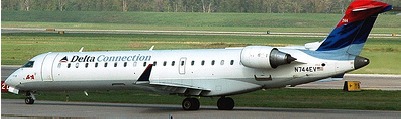}&
  \includegraphics[width=0.29\columnwidth, height = 0.22\columnwidth]{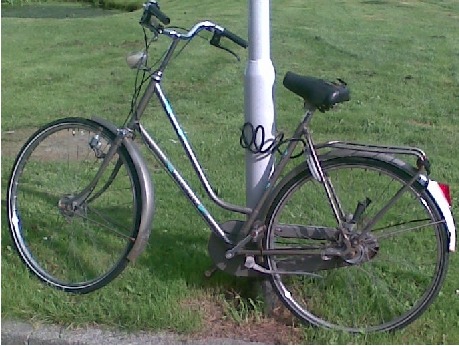}&
  \includegraphics[width=0.29\columnwidth, height = 0.22\columnwidth]{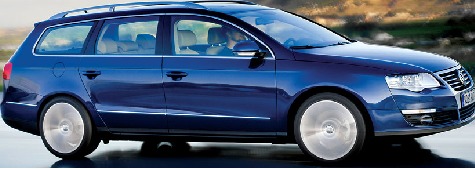}\\
  \includegraphics[width=0.29\columnwidth, height = 0.22\columnwidth]{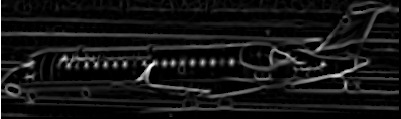}&
  \includegraphics[width=0.29\columnwidth, height = 0.22\columnwidth]{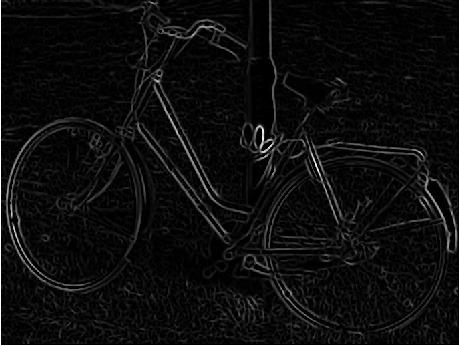}&
  \includegraphics[width=0.29\columnwidth, height = 0.22\columnwidth]{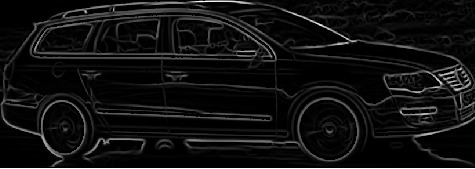}\\
  \includegraphics[width=0.29\columnwidth, height = 0.22\columnwidth]{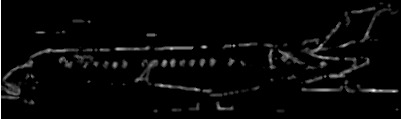}& 
  \includegraphics[width=0.29\columnwidth, height = 0.22\columnwidth]{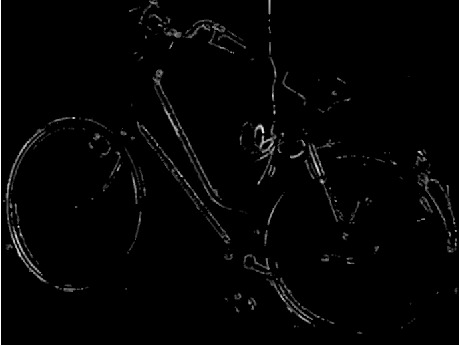}&
  \includegraphics[width=0.29\columnwidth, height = 0.22\columnwidth]{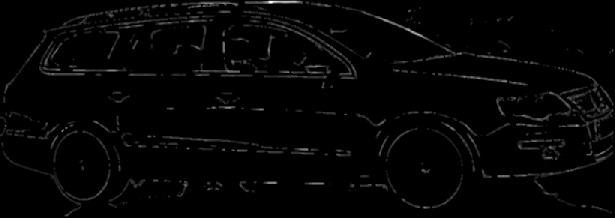}\\
  \includegraphics[width=0.29\columnwidth, height = 0.22\columnwidth]{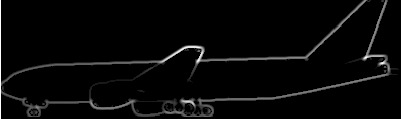}&
  \includegraphics[width=0.29\columnwidth, height = 0.22\columnwidth]{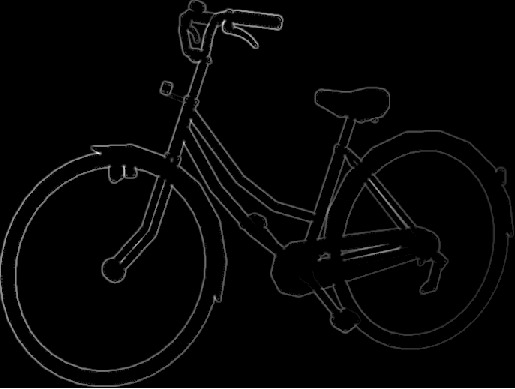}&
  \includegraphics[width=0.29\columnwidth, height = 0.22\columnwidth]{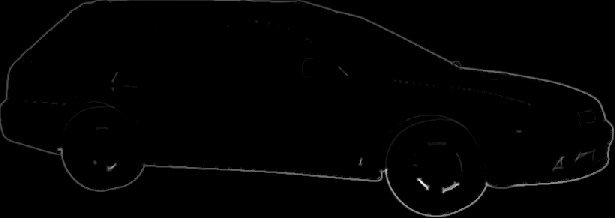}\\
  \includegraphics[width=0.29\columnwidth, height = 0.22\columnwidth]{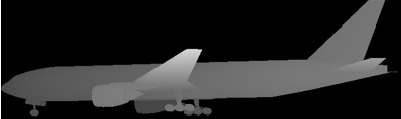}&
  \includegraphics[width=0.29\columnwidth, height = 0.22\columnwidth]{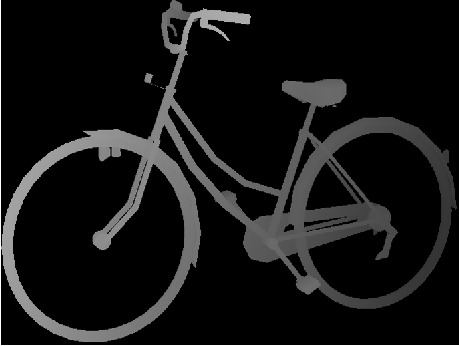}&
  \includegraphics[width=0.29\columnwidth, height = 0.22\columnwidth]{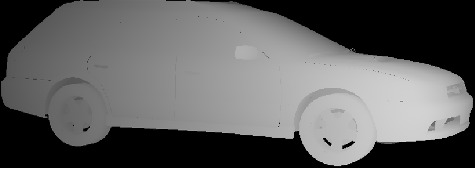}\\
\end{tabular}
  \caption{Real images (row 1), curvilinear saliency resulting with 5 scales with MCS (row~2), MFC (row~3), depth images (row~4) and CS (row~5) .}
  \label{DetectorPscal}
\end{figure}

In Fig.~\ref{DetectorPscal}, we show some visual results with the PASCAL+3D dataset. 
The MFC and MCS detectors were applied on real images and their corresponding depth images. 
As shown, MFC can reduce a lot of edges belonging to texture information and can provide an approximation of the object shapes present on depth images.

\section{Registration of 2D images to 3D models}\label{sec:registration}

In this section, we register a 2D query image to a 3D model by finding the closest view $d$ between all the rendered images of the 3D model 
${d}_k$, $k=1\ldots N$, where $N$ the number of rendered views (i.e., depth images). 
We suppose that the object to recognize is contained in a bounding box and we want to estimate the 3D pose. 
Estimating the pose consists in estimating the elevation and the azimuth angles, respectively ($h$)  and ($a$), and the distance between the model and the camera, ($v$). 
For each 3D model, we generate depth images from near uniformly distributed viewing angles around a sphere by changing $h$, $a$ and $v$ to have $N$ views per model. 
The choices for these terms are explained in paragraph~\ref{sec:results}.

To describe an object in a photograph and in all the rendered depth images, we naturally expand the famous classical HOG, Histogram Of Gradient, 
widely used in the literature~\cite{Aubry2014,Plotz2015}, to work on curvilinear saliency. 
HOG is  used in a sliding window fashion in an image to generate dense features based on binning the gradient orientation over a spatial region. 
Indeed, both in rendered depth images and in photographs, the orientation of the curvature and the magnitude of the curvilinear saliency is used for building the descriptors. 
For depth image representation, we multiply $CS$ by the eigenvector $e_{H_1}$ corresponding to the largest eigenvalue of the matrix $\mathsf{M}$ in (\ref{equ:PG:007}):
$$\overrightarrow{CS} = CS.\overrightarrow{e_{H_1}}. $$
In turn, for photographs, for $MCS$, we multiply $MCS$ values by the eigenvector $e_{S_1}$ corresponding to the curvilinear saliency $\lambda_1-\lambda_2$, 
as shown in (\ref{equ:PG:003}):
$$\overrightarrow{MCS} = MCS.\overrightarrow{e_{H_1}}. $$
In addition, for $MFC$, we also multiply $MFC$ values with the eigenvector $e_{S_1}$:
$$\overrightarrow{MFC} = MFC.\overrightarrow{e_{S_1}}. $$
Using the same principle of HOG presented in~\cite{Dalal2005}, 
we propose a descriptor that contains the orientation of the curvature and the value of curvilinear saliency and the magnitude of $CS$, $MCS$ and $MFC$ 
for an image are binned into sparse per-pixel histograms.

Given the HOG descriptor from a 2D query image $D_q$, we compute the HOG descriptors of the rendered images $D_{d_N}$, with $N$ rendering depth images. 
In order to compare $D_q$ to every $D_{d_N}$, the similarity scores are computed as proposed in~\cite{Aubry2014}:
\begin{align}
\mathbf{S}_{hog}(k,h,a,v) = (\mathbf{D}_{d_N} - \mu_s)^T\mathbf{\Sigma}^{-1} \mathbf{D_q},
\end{align}
where, $k=1...N$, $\Sigma$ and $\mu_s$ are, respectively, the covariance matrix and the mean over all descriptors of the rendered images.
For the registration process, evaluating $\mathbf{S}_{hog}(k,h,a,v)$ can be done by computing the probability of the inverse of the inner product between $D_q$ and a transformed set of descriptors. The $\mathbf{S}_{hog}(k,h,a,v)$ probability is then maximized in order to find the closest corresponding views of the query image.

We also evaluate a global similarity by measuring how well each individual detected point in an image is able to be matched with a corresponding detected point in the depth map, i.e., how well each image detected points are repeatable.
More precisely, this repeatability scores, $Rep$, normalized between 0 and 1, is the probability that key features in the intensity image are found close to those extracted in the depth image $Rep_{d_N \dashrightarrow q}$.
Since the closest view should have a high repeatability scores in comparison to other views, the dissimilarity based on repeatability scores is defined by $R_{d_i} = 1-Rep_{d_i\dashrightarrow q}.$ If we denote $\mathbf{R}_{d_i}$ the repeatability scores of $N$ rendered views of a model and a given image, the similarity $\mathbf{S}_{rep}$ is defined by:
\begin{align}
\mathbf{S}_{rep}(k,h,a,v) = exp \left(\frac{-(\mathbf{R}_{d_i}-\mu_r)^2}{2 \: \sigma_{r}^2}\right).
\end{align}
where $\mu_r$ is the mean value of $\mathbf{R}_{d_N}$ and $\sigma_{r}$ is the standard deviation (i.e., in this work $\sigma_{r} = 0.1$).

Finally, by combining all HOG feature similarities and the similarity based on the repeatability, the probability of the final similarity is given by:
\begin{align}
\mathbf{S}(m,h,a,v) =  \mathbf{S}_{hog}(k,h,a,v)\odot \mathbf{S}_{rep}(k,h,a,v).
\end{align}
where $\odot$ is the Hadamard product. Based on calculating $\mathbf{S}(k,h,a,v)$, we select at least the highest three correspondences to estimate the full pose. From the selected three views, the logically ordered or connected views (i.e., coherent views) are firstly selected.
We then find minimum and maximum values of $h$, $a$ and $v$ of the corresponding views.
Additional views are then generated in the vicinity of the selected views that is between the minimum and maximum values of the three parameters with small steps (e.g., $\delta h = 5^o$, $\delta a = 5^o$ and $\delta v = 5 cm$).
The process is again repeated for these ranges to find the closet view to the object in a query image until convergence. 
Assume the ground-truth transformation matrix ($\mathbf{T_g}$) containing rotation ($\mathbf{R_g}$) and translation ($\mathbf{t_g}$) matrices:
$\mathbf{T_g} = \left[\begin{array}{cc}
                  \mathbf{R_g} & \mathbf{t_g} \\
                  \mathbf{0}   & 1 \\
              \end{array}
              \right]$.
In addition, the estimated transformation matrix ($\mathbf{T_e}$) containing rotation ($\mathbf{R_e}$) and translation ($\mathbf{t_e}$) matrices:
$\mathbf{T_e} = \left[\begin{array}{cc}
                  \mathbf{R_e} & \mathbf{t_e} \\
                  \mathbf{0}   & 1 \\
              \end{array}
              \right]$.
Then the matrix $\mathbf{M}$ is computed: $\mathbf{M} = \mathbf{T_e}^{-1} \times \mathbf{T_g}$. The error between the two transformation matrix is: $E = \| M_i - M_j\|,$ where $i\neq j$. Thus, the convergence criterion is $E \leq \varepsilon$, $\varepsilon$ is a very small value (in this work, $\varepsilon=0.05$).

\section{Experiments for pose estimation}\label{sec:pose}

\subsection{3D models representation and alignment}\label{sec:results}

Matching photographs and rendered depth images requires a complete 3D model representation.
Each depth image represents a 3D model from different viewpoints.
Actually, we need to have a large number of depth images to completely represent a 3D model.
However, this yields a massive execution time.
Consequently, we have orthographically rendered $N$ depth images (around 700 in our experiments) from approximately uniformly distributed viewing angles $h$ and $a$ and the distance $v$
(i.e., in these experiments, $h$ is increased by a step of $50^o$, the azimuth angle, $20^o$, and the distance, $0.3$ m, for a range between $0$ and $2$ $m$).

In addition, we need a parametrization of the alignment of the model view in a depth image with the object detected in a color image.
Consequently, when we compare two models, we need to compute the optimal measure of similarity, over all possible poses.
To do this, each model is placed into a canonical coordinate frame, normalizing for translation and rotation.
Since we know the centroid of them, the models are normalized for translation by shifting them so that the center of mass is aligned with the origin.
Next, the two models are normalized for rotation by aligning the principal axes of the model with the x-, and y-axes.
It defines the ellipsoid that best fits the model. By rotating the two sets of points so that the major axis of the ellipsoid is aligned with the x-axis,
and the second major axis aligns with the y-axes, we obtain a model in a normalized coordinate frame.
We use Principle Components Analysis, PCA, to find the orientation of the major axis of the ellipse.
The set of points of the model is rotated by the difference of the direction of the two major axes.
After normalization, the two models are (near) optimally aligned and can be directly compared in their normalized poses.

In addition, in this paper, the HOG descriptor is quantized into $9$ bins, exactly as proposed in~\cite{Dalal2005}.
The photograph and each depth image is divided into a grid of square cells
(i.e., in this work, the image is divided into $8\times 8$\footnote{Different grids were tested: $4\times 4$, $8\times 8$ and $16\times 16$. The grid with $8\times 8$ size yields the best precision rate.}).
For each cell, the curvilinear saliency, focus curves or image gradients, histograms are aggregated by weighting them with their respective magnitude.

\subsection{Analysis of the results}

 \begin{figure}[htb]
  \centering
  \includegraphics[width=0.9\columnwidth]{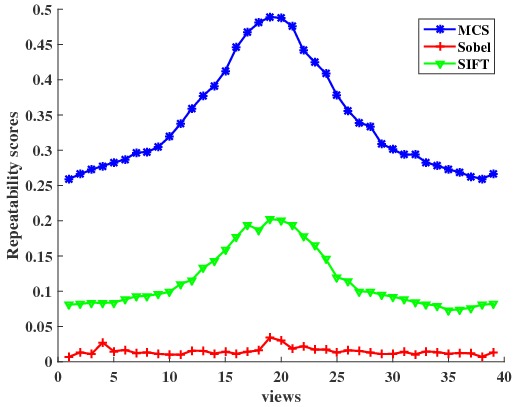}\
  \caption{Repeatability scores of an image with 40 rendered views around the correct view. The correct view is the on in the middle of the axis.}
  \label{RepScor}
\end{figure}

For pose estimation or even for object recognition, it is needed that the probability that key features in the photograph are found close to those extracted in the depth image must be high when the photograph and the depth image come from the same viewpoint.
In the first experiment that is presented, we try to illustrate that the correct view provides the highest repeatability score, in particular with the proposed detector.
In consequence, the repeatability scores of features extracted on photograph and depth images with MCS, SIFT and Sobel edge detector, 
for the correct view and the other views around it, were calculated and
shown in Fig.~\ref{RepScor} with one example of the dataset.
As shown, the three detectors yield the highest repeatability score with the correct viewpoint (even if the difference between views is small, like with Sobel).
In addition, as expected, the score is gradually diminished whenever it is far from the correct viewpoint.
The most important result is that MCS gives the highest differences between the correct view and all the other views, so, 
it illustrates that it is the most adapted detector for pose estimation based on 2D/3D registration.
This result is quite coherent because SIFT was designed to be robust against many changes, many difficulties and
it induces that the differences should be lower than MCS that is designed to be efficient in the case of 2D/3D matching.

In addition, the second experiment was performed with the Pascal+3D dataset.
For each category of objects, we compute the precision rate for detecting the correct view after using 
the three aforementioned methods for 3D model representations, i.e., Curvilinear Saliency (CS), Average Shading Gradient (ASG) and Apparent Ridges (AR)~\cite{Judd2007}, 
against the three techniques for intensity image representation, i.e., Image Gradient(IG), Multi Curvilinear Saliency (MCS) and Multi Focus Curves (MFC).
As shown in table~\ref{HDPASCAL3D2}, the registration between our Curvilinear Saliency (CS) representation of the 3D model and the multiscale focus curves
(MFC) extracted on corresponding images outperforms all other variations of the tested methods.
This confirms that curvilinear saliency representation computed from the depth images of a 3D model can capture the discontinuities of surfaces.
In addition, MFC can reduce the influence of texture and background components by extracting the edges related to the object shape in intensity images rather than MCS.
Furthermore, the precision rate is reduced by more than $25\%$ compare with ASG and IG to represent 3D models and intensity any other image representation.
Apparent ridge rendering yields the smallest registration accuracy with the three image representations among all the 3D model representation techniques.
However, using ASG with untextured 3D models against MFC and MCS also increases correct pose estimation rate.
All these results  indicate that average shading gradients computed from the normal map of an untextured geometry is a good rendering technique for the untextured geometry.
However, image gradients are not the good representation of intensity images in order to match with rendering images because it is affected by image textures. 
All these results are confirmed by the third experiment, in table~\ref{AErrorAngles}, where the details are given about the precision of the pose estimation 
in terms of elevation, azimuth, yaw angles and distance. 

\begin{table}[hbt]
\begin{center}
\setlength{\tabcolsep}{0.1cm}
\begin{tabular}{|>{\footnotesize}c|>{\footnotesize}c|>{\footnotesize}c|>{\footnotesize}c|>{\footnotesize}c|>{\footnotesize}c|>{\footnotesize}c|>{\footnotesize}c|>{\footnotesize}c|>{\footnotesize}c|>{\footnotesize}c|}
\hline
\multirow {2}{*}{Methods} &{\cellcolor{green!25}3D} &\multicolumn {3} {c|} {\cellcolor{green!25}CS}  &\multicolumn {3} {c|} {\cellcolor{green!25}ASG}  &\multicolumn {3} {c|} {\cellcolor{green!25}AR}\\ \cline{2-10}
&{\cellcolor{red!25}2D} &{\cellcolor{red!25}MFC} &{\cellcolor{red!25}MCS} &{\cellcolor{red!25}IG}  &{\cellcolor{red!25}MFC} &{\cellcolor{red!25}MCS} &{\cellcolor{red!25}IG} &{\cellcolor{red!25}MFC} &{\cellcolor{red!25}MCS} &{\cellcolor{red!25}IG}\\ \hline
\multicolumn {2}{|l|}{\cellcolor{green!25}plane}   &\cellcolor{yellow!25}{0.85}  &0.83  &0.62   &0.84 &0.80 &0.59    &0.78 &0.70 &0.50\\ \hline
\multicolumn {2}{|l|}{\cellcolor{green!25}bicycle} &\cellcolor{yellow!25}{0.81}  &0.76  &0.60   &0.80 &0.78 &0.61    &0.74 &0.73 &0.49\\ \hline
\multicolumn {2}{|l|}{\cellcolor{green!25}boat}    &\cellcolor{yellow!25}{0.78}  &0.71  &0.58   &0.75 &0.70 &0.57    &0.71 &0.68 &0.52\\ \hline
\multicolumn {2}{|l|}{\cellcolor{green!25}bus}     &\cellcolor{yellow!25}{0.87}  &0.82  &0.56   &0.82 &0.80 &0.52    &0.75 &0.74 &0.51\\ \hline
\multicolumn {2}{|l|}{\cellcolor{green!25}car}     &\cellcolor{yellow!25}{0.86}  &0.85  &0.58   &0.86 &0.83 &0.51    &0.76 &0.72 &0.47\\ \hline
\multicolumn {2}{|l|}{\cellcolor{green!25}diningtable}   &\cellcolor{yellow!25}{0.86}  &0.83  &0.61   &0.81 &0.81 &0.60    &0.79 &0.77 &0.54\\ \hline
\multicolumn {2}{|l|}{\cellcolor{green!25}motorbike}    &\cellcolor{yellow!25}{0.79}  &0.78  &0.60   &0.78 &0.75 &0.58    &0.69 &0.62 &0.52\\ \hline
\multicolumn {2}{|l|}{\cellcolor{green!25}sofa}    &\cellcolor{yellow!25}{0.85}  &0.81  &0.64   &0.80 &0.72 &0.61    &0.68 &0.61 &0.53\\ \hline
\multicolumn {2}{|l|}{\cellcolor{green!25}train}   &\cellcolor{yellow!25}{0.87}  &0.86  &0.70   &0.81 &0.82 &0.71    &0.74 &0.67 &0.58\\ \hline
\multicolumn {2}{|l|}{\cellcolor{green!25}tvmonitor}     &\cellcolor{yellow!25}{0.83}  &0.80  &0.55   &0.80 &0.79 &0.54    &0.66 &0.64 &0.52\\ \hline
\end{tabular}
\end{center}
\caption{Precision of pose estimation CS, ASG and AR against MFC, MCS and IG. }
\label{HDPASCAL3D2}
\end{table}

\begin{table}[htb]
\begin{center}
\setlength{\tabcolsep}{0.1cm}
\begin{tabular}{|>{\footnotesize}c|>{\footnotesize}c|>{\footnotesize}c|>{\footnotesize}c|>{\footnotesize}c|>{\footnotesize}c|>{\footnotesize}c|>{\footnotesize}c|>{\footnotesize}c|}
\hline
\multirow {2}{*}{Methods} &\multicolumn{2}{c|}{\cellcolor{green!25} (a)}  &\multicolumn{2}{c|}{\cellcolor{green!25} (b)}  &\multicolumn{2}{c|}{\cellcolor{green!25} (c)} &\multicolumn{2}{c|}{\cellcolor{green!25} (d)}\\ \cline{2-9}
&\cellcolor{red!25} Est.  &\cellcolor{red!25}Clo.  &\cellcolor{red!25}Est.  &\cellcolor{red!25}Clo.  &\cellcolor{red!25}Est.  &\cellcolor{red!25}Clo.  &\cellcolor{red!25}Est.  &\cellcolor{red!25}Clo. \\ \hline
\cellcolor{green!25}CS/MFC   &{$16.5^o$}  &\cellcolor{yellow!25}$4.8^o$  &\cellcolor{yellow!25}{$08.8^o$}  &\cellcolor{yellow!25}$1.2^o$  &{$5.6^o$}  &\cellcolor{yellow!25}$0.8^o$ &\cellcolor{yellow!25}18 &\cellcolor{yellow!25}7\\ \hline
\cellcolor{green!25}CS/MCS   &\cellcolor{yellow!25}{$16.0^o$} &$5.2^o$   &{$11.4^o$}  &$1.5^o$  &{$6.1^o$} &$1.1^o$  &21  &8\\ \hline
\cellcolor{green!25}ASG/MFC  &{$19.2^o$}  &$5.3^o$  &{$10.1^o$}  &$1.3^o$  &\cellcolor{yellow!25}{$5.1^o$} &\cellcolor{yellow!25}$0.8^o$ &22  &9\\ \hline
\cellcolor{green!25}ASG/MCS  &{$19.6^o$}  &$5.9^o$  &{$13.6^o$}  &$1.9^o$  &{$6.2^o$}  &$1.2^o$  &23 &11\\ \hline
\cellcolor{green!25}AR/MFC   &{$28.7^o$}  &$7.1^o$  &{$16.5^o$}  &$2.5^o$  &{$8.5^o$}  &$1.8^o$  &36  &13\\ \hline
\cellcolor{green!25}AR/MCS   &{$29.5^o$}  &$8.0^o$  &{$17.3^o$}  &$3.1^o$  &{$9.2^o$}  &$2.0^o$  &39  &17\\ \hline
\end{tabular}
\end{center}
\caption{{Average Error of the estimated (Est.) (a) elevation, (b) azimuth and (c) yaw angles and (d) distance, in centimeter, of the pose of the camera. The term Clo. indicates the closest view to the correct pose. 
These quantitative results demonstrate that the best combination is MFC/CS. }} 
\label{AErrorAngles}
\end{table}

\begin{figure*}[htb]
  \centering
\setlength{\tabcolsep}{0.1cm}
  \begin{tabular}{ccc}
  \includegraphics[width=0.65\columnwidth]{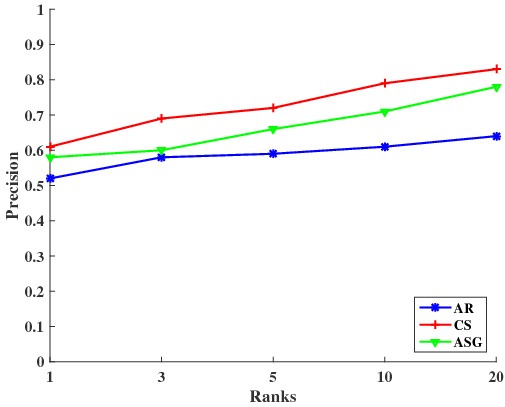}
  &\includegraphics[width=0.65\columnwidth]{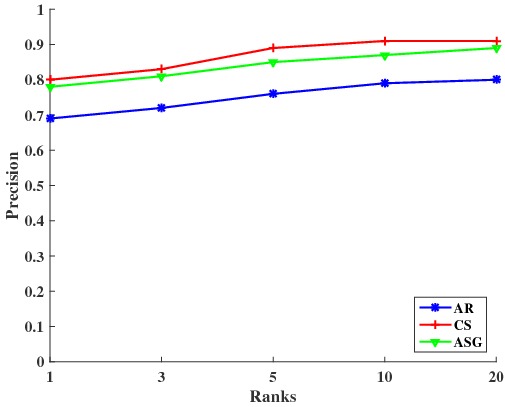}
  &\includegraphics[width=0.65\columnwidth]{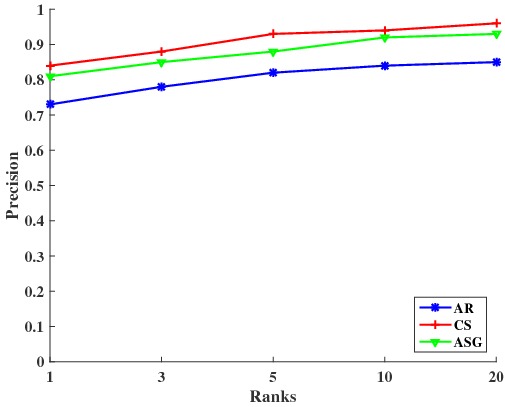}\\
  (a)&(b)&(c)\\
  \end{tabular}
  \caption{Precision values with different ranks with image representation using (a) Image Gradient (IG), (b) Multiscale Curvilinear Saliency (MCS) and (c) Multiscale Focus Curves (MFC).}
  \label{IGMCSMFCRAnks}
\end{figure*}

In the next experiment, in Fig.~\ref{IGMCSMFCRAnks}, we show the precision of the registration of images among the top $r$ similarities, i.e. 
we sort all the similarity scores obtained for all the views, and we analyse the $r$ first highest similarities (more precisely, the 1, 3, 5, 10 and 20 first ranks). 
The correct pose is searched within this set of views.
As shown, the precision rate increased when the number of views is increased, for any combination of 3D model reprensentation and image representation. 
However, MFC yields the highest precision with the three tested methods of representing 3D models (i.e., CS, ASG, and AR).
In addition, MCS yields good precision values.
In fact, IG yields the smallest precision values because the edges detected with texture information have a bad influence on estimating the successful registration.

\begin{figure*}[htb]
  \centering
  \includegraphics[width=0.6\columnwidth, height=0.45\columnwidth]{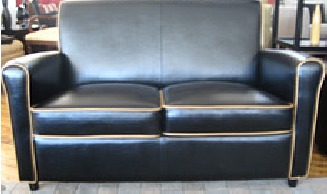}\
  \includegraphics[width=0.6\columnwidth, height=0.45\columnwidth]{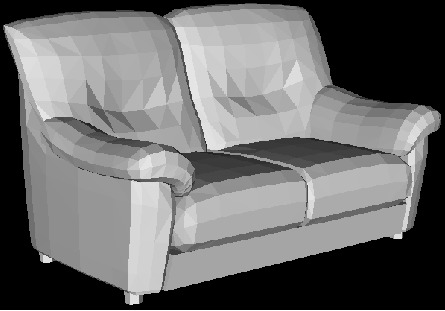}\
  \includegraphics[width=0.6\columnwidth, height=0.45\columnwidth]{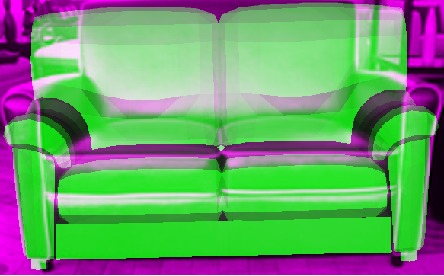}\\
  \includegraphics[width=0.6\columnwidth, height=0.45\columnwidth]{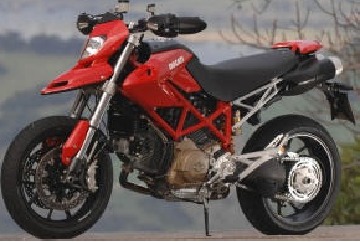}\
  \includegraphics[width=0.6\columnwidth, height=0.45\columnwidth]{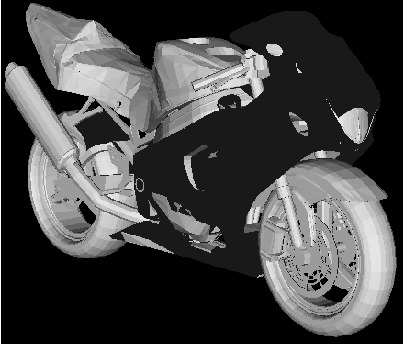}\
  \includegraphics[width=0.6\columnwidth, height=0.45\columnwidth]{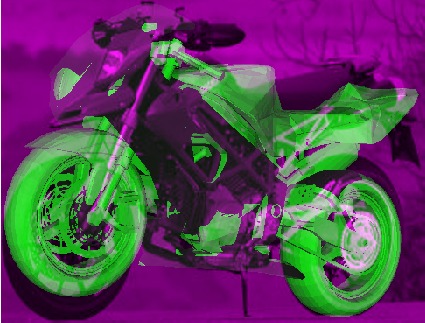}\\
  \includegraphics[width=0.6\columnwidth, height=0.45\columnwidth]{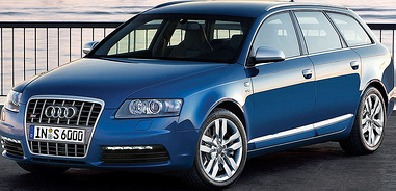}\
  \includegraphics[width=0.6\columnwidth, height=0.45\columnwidth]{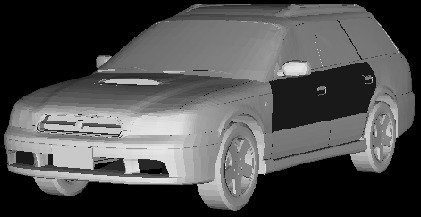}\
  \includegraphics[width=0.6\columnwidth, height=0.45\columnwidth]{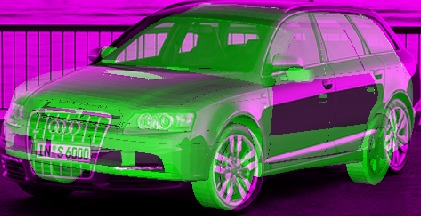}\\
  \caption{Some correct registration examples with the Pascal+3D dataset. We show the query image (column 1), the corresponding  3D model (cloumn 2) and the first ranked pose estimation (colmun 3). 
  It illustrates that even if the 3D model does not have the same detailed shape, the registration can be correctly done. }
  \label{Register}
\end{figure*}

Finally, the Fig.~\ref{Register} shows some examples of the Pascal+3D dataset of correct registrations with the top-ranked pose estimation.
It can be seen that our system is able to register an image with a great variety of texture and viewing angle.
In addition, the proposed algorithm can register images regardless of light changes in images. 

\section{Conclusion and perspective}\label{sec:conclusion}

After an analysis of existing tools for 2D/3D registration, the major goal of this paper was to propose an approach for 2D/3D matching more adapted, and in particular more justified, 
than existing approaches. For that purpose, we also proposed an evaluation protocol based on repeatibility study. 
More precisely, for doing this matching, we have studied these two important aspects: how to represent the studied data, in 2D and in 3D, and then, how to compare them. 
In this context, we introduce a 3D detector based on curvilinear saliency and a 2D detector based on the same principle but adapted in multi-scale and combined with the principle 
of focus curves. 
The interest of this new method was also illustrated by quantitative evaluation on pose estimation and 2D/3D registration. All the results are very encouraging and the next 
step of this work is to use this registration to identify defaults on objects. For that purpose, we need to study the robustness of this work to missing part of objects and 
to adapt the registration process in consequence.

\appendix

------------------------------

By computing $\mathbf{I}_x=(1,0,I_x)^\top$ and $\mathbf{I}_y=(0,1,I_y)^\top$,
 it can be easily shown that
 \begin{align}
 \mathtt{J}_\mathbf{I}^\top\mathtt{J}_\mathbf{I}
 =
 \begin{bmatrix}
 \mathbf{I}_x\cdot\mathbf{I}_x & \mathbf{I}_x\cdot\mathbf{I}_y \\
 \mathbf{I}_x\cdot\mathbf{I}_y & \mathbf{I}_y\cdot\mathbf{I}_y \\
 \end{bmatrix}
 &=
 \begin{bmatrix}
 1+(I_x)^2 & I_x I_y \\
 I_x I_y & 1+(I_y)^2  \\
 \end{bmatrix}\notag\\
 &=\mathtt{I} +\bm\nabla_I\bm\nabla_I^\top\label{equ:PG:003}
 \end{align}
---------------------------------

\end{document}